\documentclass{article}

\usepackage[final]{corl_2026}

\usepackage{graphicx}
\usepackage{amsmath}
\usepackage{amssymb}
\usepackage{comment}
\usepackage{subcaption}
\usepackage{booktabs}
\usepackage{placeins}

\title{Learning Contact Representation for Leg Odometry}

\author{
  Emre Girgin\\
  Department of Aerospace Engineering\\
  Embry Riddle Aeronautical University \\
  United States\\
  \texttt{girgine@my.erau.edu} \\
  \And
  Cagri Kilic\\
  Department of Aerospace Engineering\\
  Embry Riddle Aeronautical University \\
  United States\\
  \texttt{kilicc@erau.edu} \\
}

\begin{document}
\hypersetup{pdftitle={Learning Contact Representation for Leg Odometry}, pdfauthor={Emre Girgin}}
\maketitle

\begin{figure}[h]
    \centering
    \includegraphics[trim= 0 400 0 250, width=1\linewidth]{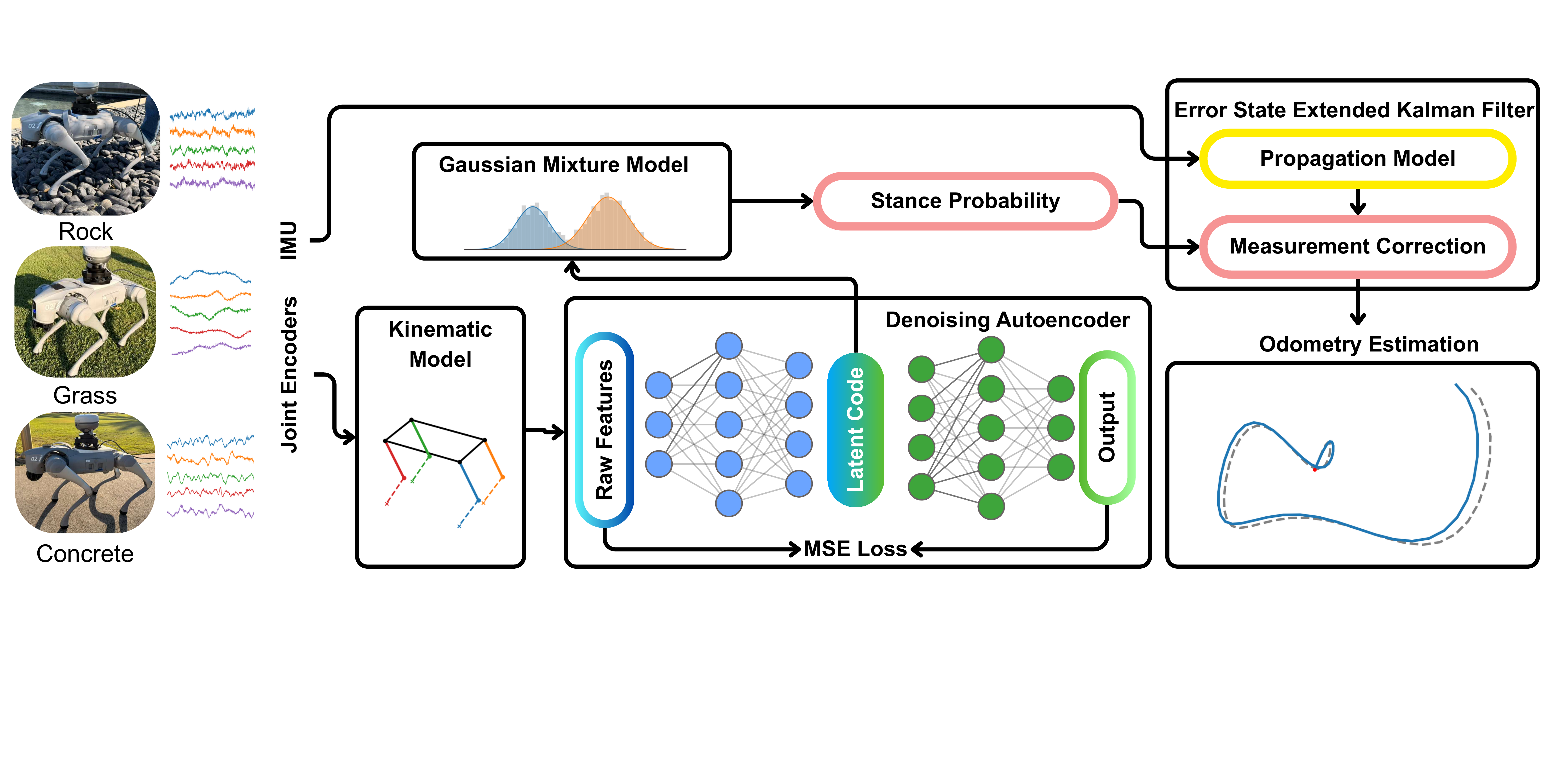}
    \caption{Our framework learns the latent representations of the leg kinematics and models the latent space as a bimodal Gaussian for the ZUPT of the ESEKF. Denoising Autoencoder is trained only on estimated foot kinematics in a self-supervised manner without relying on human supervision.}
    \label{fig:architecture}
\end{figure}

\begin{abstract}
    Leg odometry relies on the assumption that the velocity of a foot in stance is zero with respect to the world, which makes accurate contact detection a critical subproblem. Many approaches classify contact using ground reaction force (GRF) sensors at the foot tip, yet these sensors are not available on all legged robots and remain unresponsive to disturbances such as slippage while the foot is still in contact. Approaches without GRF sensing instead rely on heuristic pseudo-labels, model-based foot-velocity assumptions, or supervised contact learning. In this study, we propose a self-supervised representation learning framework for contact detection that uses only standard proprioceptive sensing (joint encoders and IMU) and requires neither force sensors nor contact labels. A denoising autoencoder is trained on kinematic features, a two-component Gaussian Mixture Model is fitted in the learned latent space, and its posterior is used as a continuous stance belief that scales the zero-velocity measurement covariance of an error-state Kalman filter. Experiments in simulation and on real-world terrain show that our framework yields the lowest odometry error among supervised and probabilistic baselines, and that higher pseudo-label classification accuracy does not translate into better downstream odometry. Our code and dataset are publicly available.\footnote{\href{https://github.com/egirgin/leg-odometry-contact-representation}{Project Repository}}
\end{abstract}


\section{Introduction}

Legged robots exhibit superior mobility in unstructured environments compared to wheeled platforms. However, their morphology complicates state estimation which is a prerequisite for downstream mapping, planning and navigation. Legged odometry relies fundamentally on the assumption that the velocity of the foot in contact with the ground is zero relative to the inertial frame \citep{hartley2020contact, bloesch2013state, 6697236, girgin2026ocelotodometrycontactestimation}, during the stance phase, referred to as Zero-Velocity Update (ZUPT) \citep{5523938, kilic2021slip}. This constraint allows kinematic chains to correct main-body velocity drift. Consequently, the accuracy of the entire downstream state estimator is bounded by the precision of the contact detection algorithm, which is a non-trivial challenge due to unmodeled contact dynamics, terrain compliance, and sensor noise.

Many existing pipelines rely on Ground Reaction Force (GRF) sensors at the foot tip to determine contact \citep{hohmeyer2025inekformer, fallon2014drift, rotella2018unsupervised, girgin2026ocelotodometrycontactestimation}. However, high-fidelity force sensors are prone to mechanical failure, suffer from shock-induced noise, and are absent on deployed minimal-sensor quadruped platforms \citep{bledt2018cheetah}. Approaches without GRF sensing instead use dynamic contact event information and foot-velocity reasoning \citep{yoon2023invariant, kim2021legged, camurri2017probabilistic, hwangbo2016probabilistic, jenelten2019dynamic} or learned contact events \citep{lin2021legged, youm2025legged}. To ensure robustness and hardware agnosticism, we propose a contact estimator framework that relies only on standard proprioceptive sensing used to obtain time series kinematic telemetry. 
To map these kinematic sequences to contact states without requiring GRF-based pseudo-labels, we adopt a self-supervised representation learning framework. A Denoising Autoencoder (DAE) projects the multivariate temporal kinematics into a latent space. Because the latent code is forced to reconstruct underlying leg dynamics, the latent representations of swing and stance appear as two dominant modes as shown in our experiments. A Gaussian Mixture Model (GMM) is then fitted to this latent space, isolating the physical contact modes into distinct probabilistic manifolds, as visualized in Figure \ref{fig:architecture}. 

Rather than a binary contact decision, our self-supervised generative approach outputs a continuous posterior probability of stance. We integrate this probability directly into an Error-State Extended Kalman Filter (ESEKF). Instead of utilizing a constant contact threshold, the continuous belief dynamically scales the measurement covariance matrix for the ZUPT, smoothing impact transients and rejecting kinematic noise during the swing phase for a better downstream leg odometry task.

We evaluate the proposed framework on an extensive, publicly available simulation dataset and on real-world data, quantifying performance through isolated contact classification and downstream leg odometry drift reduction tasks. Furthermore, ablation studies expose the nonlinear topological constraints of contact kinematics across varying temporal windows.

The main contributions of this study are: \textbf{i)} HMM-GMM, to our knowledge the first HMM contact detector for leg odometry whose emission probabilities are modeled by a GMM fitted on kinematic features, \textbf{ii)} DAE-GMM, a label-free, self-supervised framework that fits a GMM in a denoising-autoencoder latent space to extract continuous contact probabilities for leg odometry, \textbf{iii)} an ESEKF formulation whose ZUPT measurement uncertainty is driven by this latent generative belief, and \textbf{iv)} an empirical evaluation showing that contact detectors supervised on pseudo-labels or using long temporal context do not necessarily improve downstream leg odometry.

\section{Related Work}

The primary approach in leg odometry exploits the stance phase as a ZUPT \citep{bloesch2013state, 6697236, girgin2026ocelotodometrycontactestimation}. However, methods vary in their contact detection and position estimation strategies. While many studies adopt Kalman filter-based approaches \citep{hartley2020contact, girgin2026ocelotodometrycontactestimation, hartley2018legged, teng2021legged, santana2024proprioceptive, kim2025adaptive, nistico2025muse}, smoothing solutions such as factor graphs \citep{kim2021legged, yoon2023invariant} are increasingly utilized. When GRF is available via foot-mounted sensors, the simplest approach applies hard thresholding with a constant covariance \citep{hohmeyer2025inekformer, fallon2014drift}. However, this method oversimplifies complex foot-terrain interactions and neglects disturbances like slippage. Other techniques \citep{rotella2018unsupervised} model GRF as a binary clustering problem  using unsupervised methods like K-means. 

On the other hand, since GRF is not universally available, alternatives utilize joint encoder readings combined with forward kinematics. Probabilistic models estimate contact probability given kinematic data. \citet{camurri2017probabilistic} estimate stance probability via logistic regression on joint position, velocity, and torque. \citet{hwangbo2016probabilistic} employ a Hidden Markov Model (HMM) on joint readings, using Monte-Carlo sampling for transition probabilities informed by a gait pattern prior. Similarly, \citet{bledt2018contact} implement a cyclic gait scheduler to predict future contacts.~\citet{jenelten2019dynamic} also utilize an HMM, computing transition probabilities by fitting the vertical component of the estimated foot velocity to an exponential distribution to predict touchdown and liftoff events while evaluating slippage robustness. Doglegs \citep{wu2025doglegs} applies a Generalized Likelihood Ratio Test (GLRT) to foot acceleration and angular velocity, requiring sensor augmentation with foot-mounted IMUs. The OCELOT framework \citep{girgin2026ocelotodometrycontactestimation} applies a GLRT to the foot velocity, but it also relies on the GRF signal. Methods directly estimating measurement covariance require either user-defined thresholds \citep{kim2025adaptive} or fully differentiable estimators \citep{baumgartner2026coco}, coupling contact detection with the odometry estimator, limiting the approach's transferability. Recently, \citet{nistico2025muse} compare the estimated and desired foot velocities to detect slippage. 

Neural network based approaches are also gaining prominence. \citet{lin2021legged} trained a 1D CNN using kinematics for binary contact classification, generating ground truth heuristically. Another standard practice couples state estimators with locomotion policies within a reinforcement learning (RL) framework. These neural network-based state estimators are trained in simulation via supervised learning to provide observations for control policies prior to real-world transfer. \citet{ji2022concurrent} trained an MLP state estimator to supply body velocity, foot position, and contact probability observations for a concurrently trained actor-critic locomotion policy. The State Estimator Transformer (SET) \citep{yu2024state} adapts a GPT-based architecture \citep{radford2018improving}, trained alongside a control policy to facilitate agile quadrupedal actions. \citet{sun2025proprioceptive} trained a CNN-based state estimator that expands the prediction space by introducing a slip state alongside swing and stance. Finally,  \citet{youm2025legged} deployed a GRU within a similar RL framework to estimate body velocity and contact probabilities.

\section{Methodology}

In this section, we start with the odometry estimator (Section \ref{met:eseskf}) and then present the contact detectors (Section \ref{met:contact}). We first propose an HMM-GMM based probabilistic detector (Section \ref{met:contact:hmmgmm}), followed by the self-supervised neural contact detector DAE-GMM (Section \ref{met:contact:neural}). While the unsupervised HMM-GMM and the self-supervised DAE-GMM are our contributions, the supervised methods are implemented based on previous work \citep{lin2021legged, youm2025legged} as supervised baselines.

\subsection{Error-State Extended Kalman Filter}
\label{met:eseskf}

Our approach estimates the robot's odometry using an ESEKF, which decouples the system into a non-linear nominal state $\mathbf{x}$ and a linear error state $\delta\mathbf{x}$, instead of directly filtering nominal states \citep{sola2017quaternion}. This formulation minimizes linearization errors because error states evolve more linearly than nominal states, particularly for variables defined on non-linear Lie group manifolds, such as orientation. (See Appendix \ref{appx:esekf} for the full ESEKF formulation.) The nominal state $\mathbf{x}$ tracks position $\mathbf{p}^W \in \mathbb{R}^3$ and velocity $\mathbf{v}^W \in \mathbb{R}^3$ in world-frame, as well as the body to world rotation matrix $\mathbf{R} \in SO(3)$, and additive body-frame accelerometer and gyroscope biases $\mathbf{b}_a, \mathbf{b}_g \in \mathbb{R}^3$. 

The assumption that the stance foot remains static within the world frame $\{W\}$ establishes the constraint $\mathbf{v}_{\text{foot}}^W = \mathbf{0}$. The measurement function $h(\mathbf{x}_k)$ computes the predicted world-frame velocity of the stance foot by summing the base velocity $\mathbf{v}_k^W$ with the body-relative foot velocity $\mathbf{v}_{\text{rel}}^B$ at timestep $k$. Given the pseudo-measurement $\mathbf{y}_k = \mathbf{0}$, the resulting innovation $\mathbf{\nu}_k = -h(\mathbf{x}_k)$ is used in the measurement update and undergoes a Mahalanobis distance check to filter out non-stationary events like slippage \citep{6697236} or stance with high covariance. Updates are aborted if Normalized Innovation Squared (NIS) score exceeds 3-DOF Chi-Squared test at 95\% confidence level ($\gamma_{95}$).

\subsection{Contact Detectors}
\label{met:contact}

To accurately infer unobservable contact states from noisy proprioceptive sequences, we investigate probabilistic modeling with Hidden Markov Model combined with a Gaussian Mixture Model (HMM-GMM) as probabilistic detector, which is bounded by the geometric assumptions of Gaussian distributions on raw, non-linear kinematics. To resolve the boundary aliasing, we subsequently introduce deep neural networks, as supervised baselines and as our self-supervised framework, to quantify the trade-off between explicit pseudo-label supervision and probabilistic spatial separation.

While prior studies often utilize the entire input manifold (joint encoders ($\mathbf{q}$, $\mathbf{\dot{q}}$, $\mathbf{\ddot{q}}$, $\boldsymbol{\tau}$ \citep{jin2019joint}, and IMU)  as input \citep{lin2021legged}, supplying redundant or noisy features degrades generative clustering. High-dimensional inputs lead to the curse of dimensionality, resulting in ill-conditioned covariance matrices during GMM optimization and overfitting in supervised sequence models \citep{zimek2012survey}. We use a 5D subset consisting of vertical foot position relative to the base ($p_z^B$), 3D foot velocity ($\mathbf{v}_{\text{rel}}^B$) estimated by forward kinematics, and estimated calf torque ($\tau_{\text{calf}}$) for all contact detectors. We use foot velocity instead of history of foot positions to reduce the dimensionality and latency, since HMM-GMM is fitted on raw features without parallelization. (See Appendix \ref{sec:feature_selection} for details.)

\subsubsection{Gaussian Mixture Model Governed Hidden Markov Model}
\label{met:contact:hmmgmm}

We formulate legged robot contact detection as a Hidden Markov Model to estimate contact states from noisy kinematic time-series data. While prior work often focuses on dynamically adapting transition probabilities \citep{hwangbo2016probabilistic, jenelten2019dynamic}, we emphasize data-driven emission modeling.

The HMM is defined by a 5-tuple $\lambda=(\mathcal{X},\mathcal{O},B,A,\pi)$. The hidden states are $\mathcal{X} = \{\text{swing}, \text{stance}\}$ and observation space $\mathcal{O}$ consists of observation vectors $o_t \in \mathbb{R}^d$ for each time $t$ as raw features. $B$ defines the emission probabilities, $A$ represents the state transition matrix, and $\pi$ is the initial belief. To define the emission probabilities $B$, we fit an unsupervised two-component GMM to the continuous kinematic dataset, whose parameters ($\mu_j, \Sigma_j$) are optimized by Expectation Maximization. For each state $j \in \mathcal{X}$, the class-conditional emission probability is modeled as a multivariate Gaussian:
\begin{equation}
b_j(o_t)=P(o_t\mid x_t=j)=\mathcal{N}(o_t;\mu_j,\Sigma_j).
\end{equation}

The label-agnostic unsupervised clusters are mapped to physical states using kinematic constraints, where the cluster with the lower mean vertical foot position ($p_z^B$) is assigned to the stance state (see Appendix \ref{appx:cluster_assignment}). Posterior probability estimation is calculated using the forward algorithm whose details are included in Appendix \ref{appx:hmm-gmm}.

The stance belief $p_t$ is an ordinal reliability score (see Appendix \ref{appx:calibration}) and dynamically scales the measurement covariance $R_t \in \mathbb{R}^{3 \times 3}$:
\begin{equation}
R_t = \left( \frac{\alpha}{p_t + \epsilon} \right) I_{3\times3},
\end{equation}
where $\alpha=1$ is the scaling coefficient (not tuned, see Appendix \ref{appx:covariance_sensitivity}), $I$ is the identity matrix, and $\epsilon = 10^{-9}$ prevents division by zero during the swing phase and is used only for numerical stability. Lower belief produces higher covariance, leading to effective rejection of the swing leg kinematics or unreliable stances at the odometry update by the innovation based gating \citep{6697236}, without requiring hard thresholding logic.
We adopt two operational modes (Offline and Online Sliding Window) for GMM fitting to address deployment constraints. (See Appendix \ref{appx:hmm-gmm} for details.)

\begin{figure}
    \centering
    \includegraphics[trim=0 40 0 25, clip, width=0.75\linewidth]{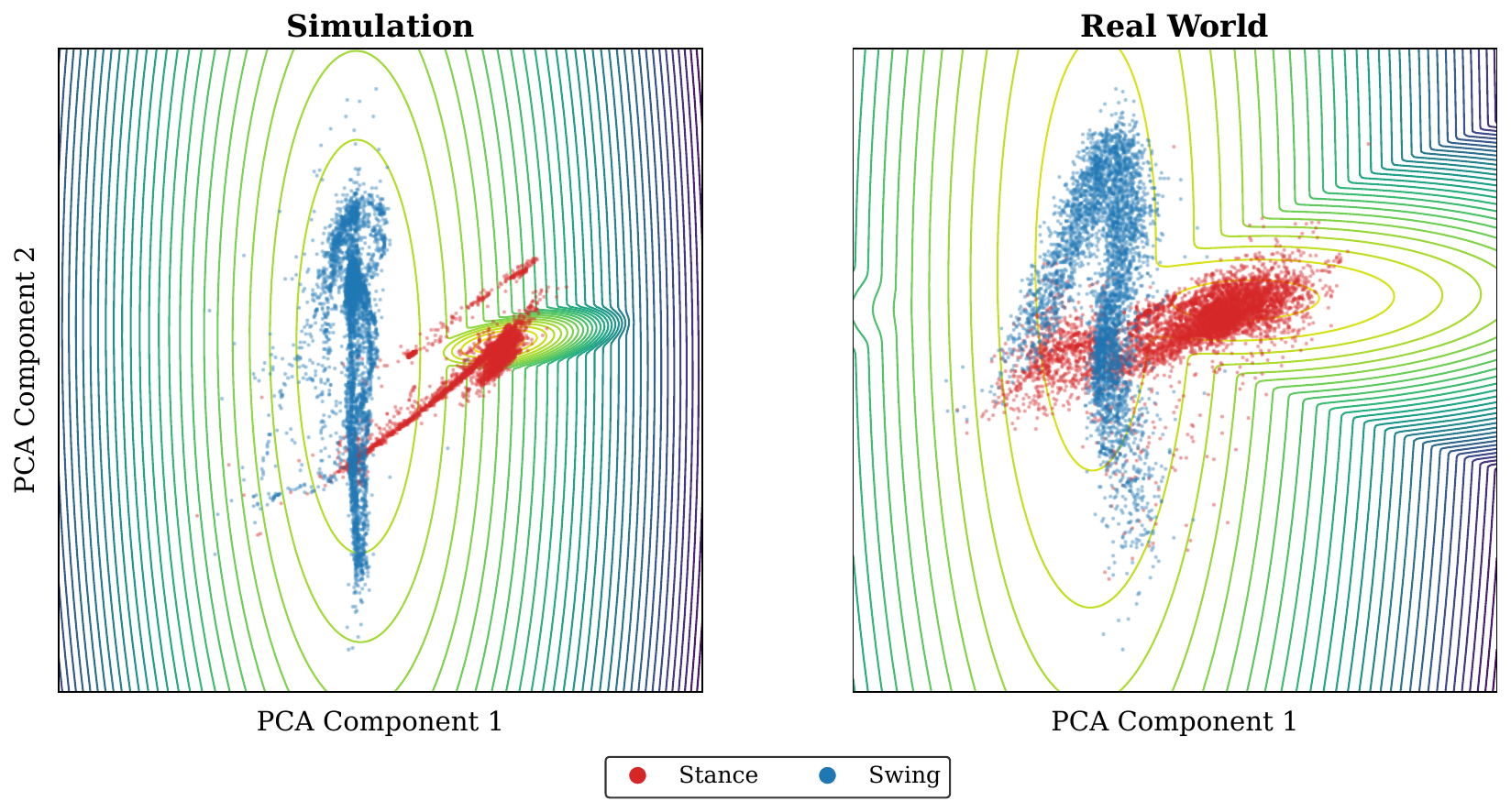}
    \caption{Visualization of input space at first two PCA components on simulation (left) and real-world sequences (right). The raw input space does not follow bimodal Gaussian distribution. Contours represent the log-likelihood levels. Red and blue colors represent the pseudo-labels.}
    \label{fig:nongaussian}
    \vspace{-0.40cm}
\end{figure}

While the HMM-GMM framework yields a continuous likelihood formulation, it is constrained by the geometric assumptions of the Gaussian distribution with convex, ellipsoidal cluster boundaries. However, the stance and swing manifolds exhibit non-linear intersections that are not naturally disjoint, particularly at transient phases such as liftoff and touchdown, which are critical for leg odometry as will be discussed in the experiments section. Figure \ref{fig:nongaussian} demonstrates the non-Gaussian nature of the raw input space. Therefore, fitting a bimodal GMM to the raw kinematic observation space results in unavoidable boundary aliasing and misclassification at the intersecting regions. The data must be projected into a latent representation space where the contact states exhibit two dominant modes (stance/swing) before fitting the generative model as in our proposed self-supervised neural framework.

\vspace{-0.3cm}

\subsubsection{Neural Contact Representation}
\label{met:contact:neural}

To address the geometric limitations of raw kinematic spaces, we utilize deep neural networks to learn features in which stance and swing are easier to separate. We evaluate both supervised direct classification as baseline and self-supervised representation learning as our proposed framework. The supervised framework leverages GRF derived pseudo-labels as targets, forcing the network to explicitly map the non-linear relationship between raw kinematics and binary contact states. Conversely, the self-supervised framework avoids label dependency entirely. It focuses purely on representation learning, projecting the raw kinematics into a latent space designed to satisfy the geometric assumptions of the Gaussian formulation.

\noindent \textbf{Supervised Neural Contact Detectors:}
For our supervised baselines, we formulate the detection problem as a dense time-series classification task. Prior to sequence extraction, the raw observation vectors $o_t$ are normalized using the training set statistics. At each time step $t$, for an input vector of dimension $C$ and window size $W$, the sequence is defined as $O_{t} = [o_{t-W+1}, \dots, o_{t}] \in \mathbb{R}^{C \times W}$. This causal formulation ensures no future data is utilized, maintaining compatibility with real-time, online deployment constraints. Both networks output a single logit, which is mapped to a stance probability $\hat{p}_t \in (0, 1)$ via a sigmoid activation function. We use two architectures for supervised contact detection: a 1D Convolutional Neural Network (CNN) from \citet{lin2021legged} and a Gated Recurrent Unit (GRU) network from \citet{youm2025legged}. (See Appendix \ref{appx:supervised} for architectural details.)

For contact targets, we generate binary pseudo-labels $\tilde{y}_t \in \{0, 1\}$ from GRF data obtained during training data collection, depending on the noise characteristics of the source domain. In simulation, GRF signals ($F_z$) exhibit negligible noise. We utilize a strict thresholding mechanism: $\tilde{y}_t = 1$ if $F_z > \mathbb{T}_{\text{sim}}$, otherwise $0$. On the other hand, in the real world, the physical robot telemetry exhibits significant sensor noise, making hard thresholding unstable. We fit a continuous HMM-GMM explicitly to the 1D $F_z$ signal. The discrete contact state is extracted from the normalized posterior belief to provide a robust pseudo-label which is immune to transient force spikes. Both networks are optimized via standard Binary Cross-Entropy (BCE) loss.

\noindent \textbf{Self-Supervised Neural Contact Detectors:} While supervised sequential models provide discriminative decision boundaries, their reliance on GRF pseudo-labels presents a critical vulnerability. To eliminate label dependency while retaining a probabilistically continuous contact belief, we propose a Denoising Autoencoder (DAE) framework coupled with a latent-space GMM.

The objective of the DAE is to project the entangled kinematic window into a latent space $\mathcal{Z}$ where the underlying physical states (stance vs. swing) form two dominant modes under an encoder network $f_\theta$ (see Figure \ref{fig:window_size_analysis} in Section \ref{sec:experiments}). The standardized kinematic sequence $O_t \in \mathbb{R}^{C \times W}$ is explicitly augmented via Gaussian noise ($\mathcal{N}(0, \sigma^2)$) and uniform amplitude scaling to prevent the network from learning a trivial identity function. We evaluate both CNN and GRU architectures for $f_\theta$. (See Appendix \ref{appx:self-supervised} for architectural details.) Even though our design supports varying window sizes, we choose $W=1$ by default, effectively reducing CNN architecture to MLP and GRU to a nonlinear transform. However, we keep the window size in our architecture to demonstrate the effect of varying window sizes in our ablation study. A decoder network $g_\phi$, comprised of 1D Transposed Convolutions, reconstructs the uncorrupted original sequence from the latent code: $\hat{O}_t = g_\phi(z_t)$. The network is optimized via Mean Squared Error (MSE) reconstruction loss.

By forcing the DAE to reconstruct the underlying dynamics of the leg, the latent code $z_t$ intrinsically encodes the contact mode, whose size is found as $D=16$ empirically (see Appendix \ref{appx:latent_dim}), so that stance and swing form two dominant modes in $\mathbb{R}^D$. To extract a continuous contact belief, we fit an unsupervised 2-component GMM to the latent representations $\mathcal{Z}$ which is obtained from the training set. Note that the GMM is fitted to the training data during DAE training and those precomputed probability distributions are used for inference to avoid non-parallelized fitting of GMM in real-time.
Since clusters lack explicit labels, we assign physical states using the identical heuristic defined in our HMM formulation. The cluster mapping to the samples with the lowest mean vertical foot position ($p_z^B$) is designated as the stance component.

\section{Experiments}
\label{sec:experiments}

All contact detection frameworks are evaluated across their classification performance and downstream impact on quadruped leg odometry. We further conduct ablation studies to reveal intrinsic properties of the contact kinematics. For the experiments, TartanGround \citep{patel2025tartanground} simulation dataset and our real-world dataset on diverse terrain such as concrete, grass and rock are utilized. The models are trained per dataset so a single model is shared across all terrains of a dataset. (See Appendix \ref{appx:experiments} for details on datasets, default hyperparameter set and additional experiments.) 

We benchmark the following contact estimation pipelines: \textbf{i)} GRF-Thresholding (Baseline): Hard thresholding on $F_z$. \textbf{ii)} HMM-GMM (Offline and Online): Our unsupervised probabilistic detector, which models continuous probability on raw kinematic inputs. \textbf{iii)} Supervised NN (CNN \citep{lin2021legged} and GRU \citep{youm2025legged}): Discriminative models trained on GRF pseudo-labels. \textbf{iv)} Self-supervised DAE-GMM (CNN and GRU): The proposed label-free representation learning pipelines. Every contact detector scales the ZUPT covariance in the same way and only the source of $p_t$ differs: the sigmoid output for the supervised networks, $F_z/\mathbb{T}$ clamped to $[0,1]$ for GRF thresholding, and the posterior probability for HMM-GMM and DAE-GMM.

\subsection{Classification Accuracy and Domain Shift}

Standard evaluation protocols for leg odometry contact detection rely on classification metrics against heuristic pseudo-labels \citep{lin2021legged, sun2025proprioceptive}. We establish this comparison by evaluating isolated contact classification performance, detailed in Table \ref{tab:classification_metrics}. Supervised CNN and GRU models achieve the highest scores on both datasets. However, all detectors exhibit performance degradation when evaluated on the physical robot due to sensor noise and unmodeled terrain compliance in the real world. 

\begin{table}[h!]
\centering
\caption{Contact Classification Metrics Across Domains $\uparrow$ (\textbf{Best}, \underline{Second Best})}
\label{tab:classification_metrics}
\begin{tabular}{l|ccc|ccc}
\toprule
& \multicolumn{3}{c|}{TartanGround (Sim)} & \multicolumn{3}{c}{Real-World (Physical)} \\
\textbf{Method} & \textbf{Precision} & \textbf{Recall} & \textbf{F1} & \textbf{Precision} & \textbf{Recall} & \textbf{F1} \\
\midrule
Supervised CNN \citep{lin2021legged}    & \underline{0.996} & \textbf{0.999} & \textbf{0.998} & \textbf{0.935} & 0.895 & \underline{0.907} \\
Supervised GRU \citep{youm2025legged}   & \textbf{0.997} & \textbf{0.999} & \textbf{0.998} & \underline{0.924} & 0.929 & \textbf{0.921} \\
\midrule
HMM-GMM (Offline) & 0.995 & 0.980 & 0.988 & 0.916 & 0.922 & 0.906 \\
HMM-GMM (Online)  & 0.994 & 0.937 & 0.964 & 0.917 & 0.868 & 0.886 \\
DAE-GMM (CNN)  & 0.983 & \underline{0.994} & 0.988 & 0.809 & \underline{0.967} & 0.868 \\
DAE-GMM (GRU)  & 0.985 & 0.992 & \underline{0.989} & 0.844 & \textbf{0.994} & 0.900 \\
\bottomrule
\end{tabular}
\end{table}

\begin{figure}[t]
    \centering
    \includegraphics[trim= 0 18 0 18, width=1\linewidth]{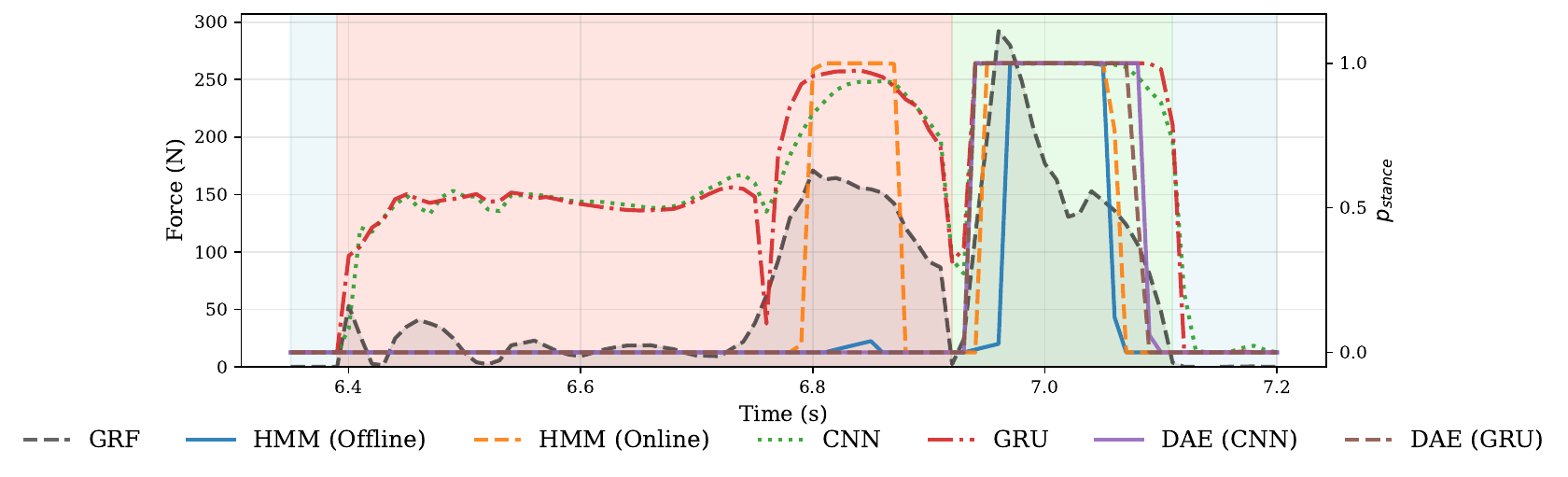}
    \caption{Stance probabilities across contact detectors during a slippage event, with ground-truth (GT) GRF shown for reference (grey dotted). The non-zero GRF wrongfully leads supervised methods to generate high probability of stance and to overestimate the stance duration, during unstable contact (light red background). The proposed DAE-GMM methods isolate the reliable stance period (light green background).}
    \label{fig:classification_plot}
\end{figure}

Crucially, ground truth pseudo-labels rely on heuristics. Table \ref{tab:classification_metrics} demonstrates how supervised models successfully fit to these heuristics. However, we hypothesize that heuristic pseudo-labels are fundamentally inadequate for identifying the physically reliable stance of the leg during dynamic locomotion. Figure \ref{fig:classification_plot} demonstrates an example GRF signal and stance probabilities estimated by contact detectors, collected during an example slippage period. While supervised models generate higher probability for the section where the contact is clearly weak and the robot is not standing on that leg, our self-supervised models produce zero probability of stance until the robot shifts its weight to that leg. The overestimation of the stance duration by supervised methods leads to overconfident measurement update and hurts filter accuracy. Consequently, in the proposed framework, we formulate contact estimation as a self-supervised continuous stance probability problem, making it more robust to slippage and noise at GRF sensor. Offline HMM-GMM follows a trend similar to DAE-GMM, but it underestimates the contact period preventing maximum correction in ESEKF. The following section demonstrates that superior classification performance on pseudo-labels does not correlate with downstream odometry accuracy.

\subsection{Downstream Leg Odometry Performance}

We evaluate the practical utility of continuous contact estimation via closed-loop ESEKF trajectory estimation. Trajectories are evaluated using RMSE for time-synchronized absolute errors (Absolute Trajectory Error-ATE, Absolute Heading Error-AHE), relative errors (Relative Position Error-RPE translation/rotation) \citep{sturm2012benchmark}, and time-synchronization-free geometric errors (Final Position Error-FPE, Fréchet Distance). (See Appendix \ref{sec:metrics} for details.)

Table \ref{tab:tartanground_contact_detector_comparison} reports odometry errors on the TartanGround dataset. The proposed DAE-GMM (CNN) and DAE-GMM (GRU) methods achieve the lowest errors across all metrics, despite supervised methods demonstrating higher pseudo-label classification accuracy (Table \ref{tab:classification_metrics}). Furthermore, supervised models underperform the GRF thresholding baseline. This indicates that the generalization capabilities of neural networks do not compensate for the noise injected during heuristic pseudo-labeling.

\begin{table}[h]
\centering
\caption{Leg Odometry Performance on the Simulation Dataset $\downarrow$  (\textbf{Best}, \underline{Second Best})}
\label{tab:tartanground_contact_detector_comparison}
\small
\begin{tabular}{lcccccc}
\toprule
\textbf{Contact Detector} & \textbf{ATE} & \textbf{AHE} & \textbf{RPE Trans} & \textbf{RPE Rot} & \textbf{FPE} & \textbf{Fréchet} \\ 
 & \textbf{(m)} & \textbf{($^\circ$)} & \textbf{(\%)} & \textbf{($^\circ$/m)} & \textbf{(m)} & \textbf{(m)} \\
\midrule
GRF Threshold    & 4.25 & 3.03 & 10.20 & 0.143 & 7.48 & 7.64 \\
Supervised CNN \citep{lin2021legged}      & 4.32 & 3.27 & 10.45 & 0.147 & 7.64 & 7.80 \\
Supervised GRU \citep{youm2025legged}     & 4.29 & 3.14 & 10.33 & 0.146 & 7.54 & 7.68 \\ 
\midrule
HMM-GMM (Offline)     & 4.50 & 3.45 & 10.72 & 0.146 & 8.13 & 8.28 \\
HMM-GMM (Online)      & 4.88 & 3.86 & 12.41 & 0.181 & 9.07 & 9.25 \\ 
\midrule
DAE-GMM (CNN)     & \textbf{4.09} & \textbf{2.69} & \textbf{9.93} & \textbf{0.141} & \textbf{7.13} & \textbf{7.28} \\
DAE-GMM (GRU)     & \underline{4.15} & \underline{2.81} & \underline{10.05} & \underline{0.142} & \underline{7.25} & \underline{7.44} \\ 
\bottomrule
\end{tabular}
\end{table}

Table \ref{tab:ocelot_contact_detector_comparison} details real-world odometry performance. The performance gap between the proposed DAE-GMMs and the baseline methods widens on physical hardware. DAE-GMM (CNN) yields the lowest errors in all metrics, while DAE-GMM (GRU) is the second best at every metric except rotational RPE. Because real-world ground truth relies on an offline 1D GRF HMM-GMM heuristic, supervised networks trained on this data outperform GRF thresholding, reversing the trend observed in simulation.

\begin{table}[h]
\centering
\caption{Leg Odometry Performance on the Real-World Dataset $\downarrow$  (\textbf{Best}, \underline{Second Best})}
\label{tab:ocelot_contact_detector_comparison}
\small
\begin{tabular}{lcccccc}
\toprule
\textbf{Contact Detector} & \textbf{ATE} & \textbf{AHE} & \textbf{RPE Trans} & \textbf{RPE Rot} & \textbf{FPE} & \textbf{Fréchet} \\ 
 & \textbf{(m)} & \textbf{($^\circ$)} & \textbf{(\%)} & \textbf{($^\circ$/m)} & \textbf{(m)} & \textbf{(m)} \\
\midrule
GRF Threshold       & 25.36 & 55.12 & 76.21 & 3.33 & 57.24 & 57.07 \\
Supervised CNN \citep{lin2021legged}          & 21.95 & 44.67 & 68.42 & \underline{3.24} & 50.95 & 50.76 \\
Supervised GRU  \citep{youm2025legged}          & 22.16 & 45.45 & 69.37 & 3.25 & 51.47 & 51.37 \\ 
\midrule
HMM-GMM (Offline)      & 27.13 & 57.13 & 82.98 & 3.36 & 61.82 & 61.76 \\
HMM-GMM (Online)      & 28.68 & 61.18 & 88.72 & 3.33 & 64.84 & 64.72 \\ 
\midrule
DAE-GMM (CNN)             & \textbf{17.20} & \textbf{31.90} & \textbf{49.93} & \textbf{3.22} & \textbf{28.66} & \textbf{30.54} \\
DAE-GMM (GRU)             & \underline{18.87} & \underline{44.11} & \underline{62.08} & 3.30 & \underline{40.07} & \underline{39.95} \\ 
\bottomrule
\end{tabular}
\end{table}

The HMM-GMM variants yield the highest errors overall, though they remain competitive, especially in rotational RPE. The online sliding-window variant marginally underperforms the offline model, consistent with the reduced context.

To summarize, supervised models are bounded by the accuracy of the pseudo-labeling method, whereas the unsupervised HMM-GMM method is limited due to non-Gaussian nature of the raw input space. Overall, the leg odometry experiments validate that formulating contact detection as a self-supervised continuous probability density estimation task rather than supervised binary classification is a more robust approach. After all, binary contact classification is fundamentally ill-posed. Even given noise-free GRF signals, unmodeled foot-terrain compliance makes binary states inadequate for high-accuracy state estimation during dynamic locomotion.

\subsection{Ablation Studies and Analyses}

\begin{figure}[htbp]
    \centering
    \begin{subfigure}[b]{0.18\textwidth}
        \centering
        \includegraphics[trim= 0 50 0 45, width=\textwidth]{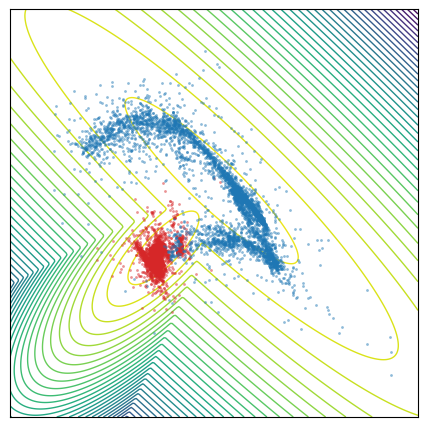}
    \end{subfigure}
    \hfill
    \begin{subfigure}[b]{0.18\textwidth}
        \centering
        \reflectbox{\includegraphics[trim= 0 45 0 50, width=\textwidth, angle=180,  origin=c]{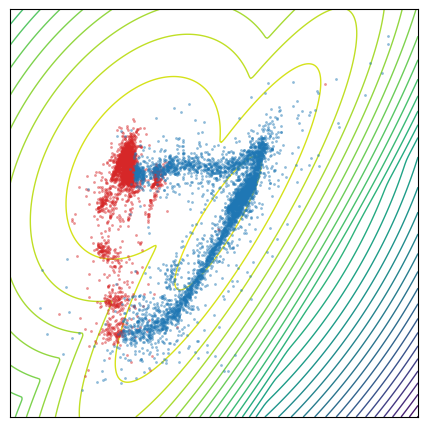}}
    \end{subfigure}
    \hfill
    \begin{subfigure}[b]{0.18\textwidth}
        \centering
        \includegraphics[trim= 0 50 0 45, width=\textwidth]{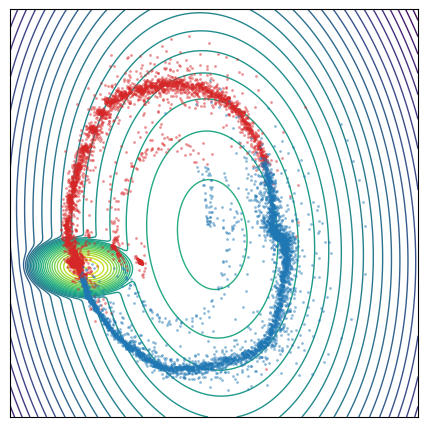}
    \end{subfigure}
    \hfill
    \begin{subfigure}[b]{0.18\textwidth}
        \centering
        \includegraphics[trim= 0 50 0 45, width=\textwidth]{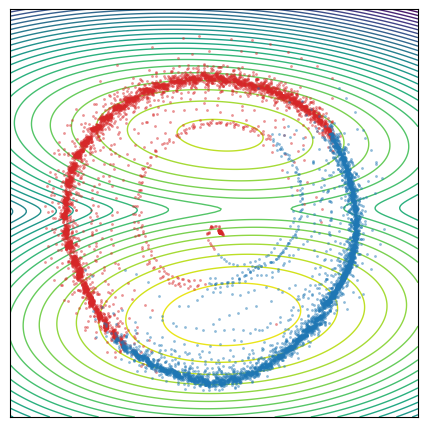}
    \end{subfigure}
    \hfill
    \begin{subfigure}[b]{0.18\textwidth}
        \centering
        \includegraphics[trim= 0 45 0 50, width=\textwidth, angle=180, origin=c]{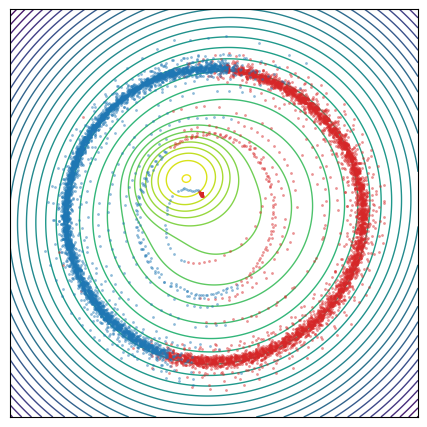}
    \end{subfigure}
    
    \vspace{1em}

    \begin{subfigure}[b]{0.18\textwidth}
        \centering
        \includegraphics[width=\textwidth]{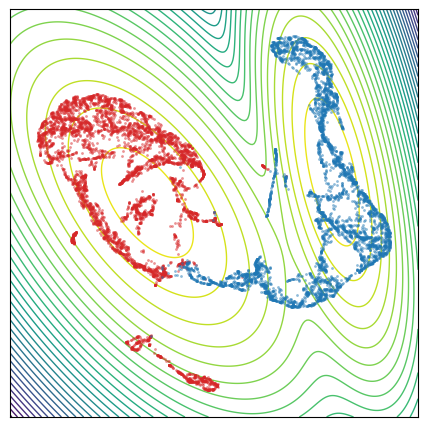}
        \caption{$W= 1$}
    \end{subfigure}
    \hfill
    \begin{subfigure}[b]{0.18\textwidth}
        \centering
        \includegraphics[width=\textwidth, angle=180,  origin=c]{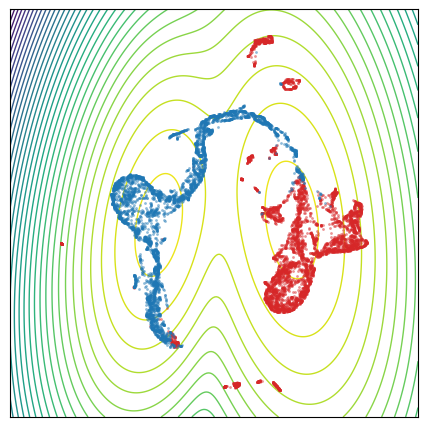}
        \caption{$W= 5$}
    \end{subfigure}
    \hfill
    \begin{subfigure}[b]{0.18\textwidth}
        \centering
        \includegraphics[width=\textwidth]{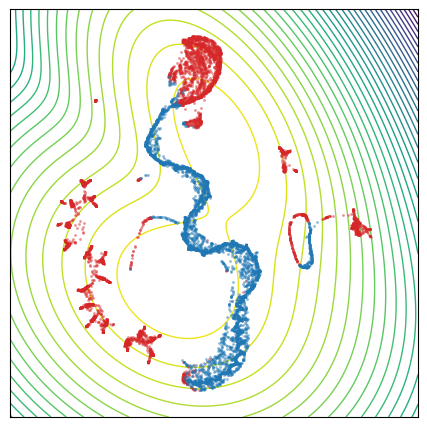}
        \caption{$W= 20$}
    \end{subfigure}
    \hfill
    \begin{subfigure}[b]{0.18\textwidth}
        \centering
        \includegraphics[width=\textwidth]{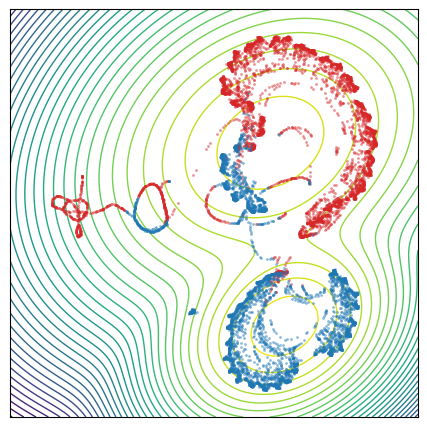}
        \caption{$W= 50$}
    \end{subfigure}
    \hfill
    \begin{subfigure}[b]{0.18\textwidth}
        \centering
        \reflectbox{\includegraphics[width=\textwidth]{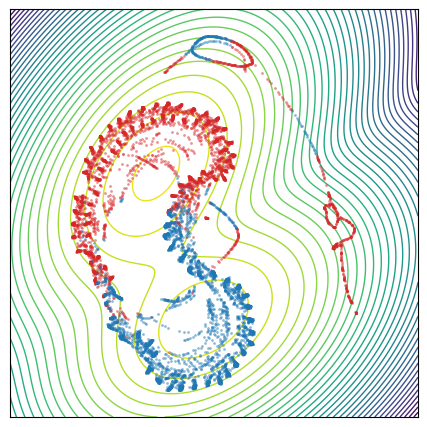}}
        \caption{$W= 100$}
    \end{subfigure}
    
    \caption{Latent space projected onto PCA (top) and UMAP (bottom) dimensions. Colors indicate pseudo contact labels (red for stance, blue for swing). Contours represent the log-likelihood of the fitted GMM. As window size increases, cyclic gait patterns dominate the geometry, forming a nonlinear ring manifold. The overlap at the boundary corresponds to transitions (touchdown/liftoff) and gets an intermediate posterior.}
    \label{fig:window_size_analysis}
\end{figure}

\noindent \textbf{Window Size and Temporal History:} 
While time-series kinematic data typically benefits from historical context, in our experiment, increasing the input window size causes the periodic gait to dominate the latent space. This forms a nonlinear, ring-like manifold, prominently visible in the PCA projections (Figure \ref{fig:window_size_analysis}, top row). Standard GMMs fail on this topology because their covariance ellipsoids assume a Euclidean space. Therefore, we choose window size $W=1$ by default for both CNN and GRU based autoencoders. Even though periodic gait is not modeled in our approach, these periodic features may be useful when modeled as prior in a Bayesian framework \citep{hwangbo2016probabilistic}. To evaluate a nonlinear mapping, we applied a 2-component UMAP \citep{mcinnes2018umap} dimensionality reduction to the latent space prior to GMM fitting. While UMAP allows GMM clustering on larger window sizes, the ring-like distribution of the periodic gait remains unavoidable (Figure \ref{fig:window_size_analysis}, bottom row), which degrades contact state separation. 
Consequently, minimizing the history window is required to preserve a latent space with two dominant, Gaussian-like modes suitable for real-time GMM inference.

\noindent \textbf{Reliability of the Stance Posterior:} Against GRF pseudo-labels, the DAE-GMM posterior is not calibrated (ECE $=0.080$), but it is a strong ordinal reliability score: its Spearman correlation with the ZUPT innovation magnitude is $\rho=-0.408$ (GRU) and $\rho=-0.380$ (CNN), compared to $\rho=-0.029$ for HMM-GMM (Offline). Higher stance scores thus consistently coincide with smaller innovations (Appendix \ref{appx:calibration}).

\noindent \textbf{Covariance Scaling:} Sweeping the scaling coefficient over $\alpha \in [0.25, 2.0]$ shows that the untuned $\alpha=1$ lies in a well-behaved region, with $\alpha$ mainly affecting yaw drift (Appendix \ref{appx:covariance_sensitivity}).

\noindent \textbf{Number of GMM Components:} With 3 to 10 components, the additional components fall into the transition zone between stance and swing and mostly capture touchdown rather than a consistent slip mode. Two components therefore suffice for ZUPT, with ambiguous contacts represented by an intermediate posterior (Appendix \ref{appx:num_components}).

\noindent \textbf{Elevation Changes:} On upward and downward staircase sequences, stance samples can occur at high foot elevations, yet the stance cluster retains a lower mean foot z-position than the swing cluster, so the cluster-to-state assignment remains valid (Appendix \ref{appx:cluster_assignment}).

\noindent \textbf{Computational Latency:} 
To benchmark computational requirements, we measured per-step inference latency on a desktop workstation (Intel i9-14900K CPU, NVIDIA RTX 4090 GPU). These baselines are computed without model quantization or hardware-specific acceleration. The GRF thresholding baseline executes fastest at $<1.0$ ms. The kinematic HMM-GMM requires 1.3 ms for online processing and 3.0 ms offline. The proposed DAE-GMMs (CNN and GRU) execute in 3.0 ms, slightly faster than their supervised counterparts at 3.5 ms. All evaluated methods remain bounded within the 10 ms execution budget dictated by 100 Hz odometry control loop.

\section{Limitations}

Although our self-supervised contact estimator improves over alternative approaches, several limitations remain. First, the latent mapping discards the input sensor noise, so the bimodal Gaussian model assumes that latent distributions are independent of input noise. Second, although we adopt an isotropic covariance, contact slippage can be highly anisotropic. Third, the posterior is an ordinal reliability score rather than a calibrated probability. Replacing the deterministic DAE with a Variational Autoencoder could formally parameterize the latent space as a probability distribution. Fourth, slip, weak contact, and partial loading are represented only through an intermediate posterior rather than as explicit contact states, and the cluster-to-state heuristic was validated on staircases but not on rubble, stepping stones, or steep slopes.

\section{Conclusion}

This study proposes a self-supervised representation learning framework for contact modeling in legged odometry that requires neither force sensors nor contact labels. Evaluations against supervised and probabilistic unsupervised baselines in simulation and on hardware show that ZUPT reduces state estimation drift when its covariance is scaled by the posterior of a GMM fitted in the DAE latent space, where stance and swing appear as two dominant modes, rather than in the raw kinematic space. We further show that higher pseudo-label classification accuracy does not necessarily translate into better leg odometry. Finally, longer temporal context biases the latent representation toward periodic gait dynamics, which could serve as priors within Bayesian filtering architectures.

\clearpage
\acknowledgments{We would like to thank the reviewers and the area chair for their valuable comments. Also, we thank Tuba Girgin for her support and constructive reviews.}

\bibliography{example}

@article{bloesch2013state,
  title={State estimation for legged robots -- consistent fusion of leg kinematics and {IMU}},
  author={Bloesch, Michael and Hutter, Marco and Hoepflinger, Mark A and Leutenegger, Stefan and Gehring, Christian and Remy, C David and Siegwart, Roland},
  journal={Robotics},
  volume={17},
  pages={17--24},
  year={2013},
  publisher={The MIT Press}
}

@INPROCEEDINGS{6697236,
  author={Bloesch, Michael and Gehring, Christian and Fankhauser, Péter and Hutter, Marco and Hoepflinger, Mark A. and Siegwart, Roland},
  booktitle={2013 IEEE/RSJ International Conference on Intelligent Robots and Systems}, 
  title={State estimation for legged robots on unstable and slippery terrain}, 
  year={2013},
  volume={},
  number={},
  pages={6058-6064},
  doi={10.1109/IROS.2013.6697236}}

@article{hartley2020contact,
  title={Contact-aided invariant extended {Kalman} filtering for robot state estimation},
  author={Hartley, Ross and Ghaffari, Maani and Eustice, Ryan M and Grizzle, Jessy W},
  journal={The International Journal of Robotics Research},
  volume={39},
  number={4},
  pages={402--430},
  year={2020},
  publisher={Sage Publications Sage UK: London, England}
}

@ARTICLE{5523938,
  author={Skog, Isaac and Handel, Peter and Nilsson, John-Olof and Rantakokko, Jouni},
  journal={IEEE Transactions on Biomedical Engineering}, 
  title={Zero-Velocity Detection—An Algorithm Evaluation}, 
  year={2010},
  volume={57},
  number={11},
  pages={2657-2666},
  doi={10.1109/TBME.2010.2060723}}

@article{kilic2021slip,
  title={Slip-based autonomous {ZUPT} through {Gaussian} process to improve planetary rover localization},
  author={Kilic, Cagri and Ohi, Nicholas and Gu, Yu and Gross, Jason N},
  journal={IEEE Robotics and Automation Letters},
  volume={6},
  number={3},
  pages={4782--4789},
  year={2021},
  publisher={IEEE}
}

@inproceedings{bledt2018cheetah,
  title={{MIT} {Cheetah} 3: Design and control of a robust, dynamic quadruped robot},
  author={Bledt, Gerardo and Powell, Matthew J and Katz, Benjamin and Di Carlo, Jared and Wensing, Patrick M and Kim, Sangbae},
  booktitle={2018 IEEE/RSJ International Conference on Intelligent Robots and Systems (IROS)},
  pages={2245--2252},
  year={2018},
  organization={IEEE}
}

@inproceedings{hartley2018legged,
  title={Legged robot state-estimation through combined forward kinematic and preintegrated contact factors},
  author={Hartley, Ross and Mangelson, Josh and Gan, Lu and Jadidi, Maani Ghaffari and Walls, Jeffrey M and Eustice, Ryan M and Grizzle, Jessy W},
  booktitle={2018 IEEE International Conference on Robotics and Automation (ICRA)},
  pages={4422--4429},
  year={2018},
  organization={IEEE}
}

@inproceedings{teng2021legged,
  title={Legged robot state estimation in slippery environments using invariant extended {Kalman} filter with velocity update},
  author={Teng, Sangli and Mueller, Mark Wilfried and Sreenath, Koushil},
  booktitle={2021 IEEE International Conference on Robotics and Automation (ICRA)},
  pages={3104--3110},
  year={2021},
  organization={IEEE}
}

@inproceedings{santana2024proprioceptive,
  title={Proprioceptive state estimation for quadruped robots using invariant {Kalman} filtering and scale-variant robust cost functions},
  author={Santana, Hilton Marques Souza and Soares, Jo{\~a}o Carlos Virgolino and Nistic{\`o}, Ylenia and Meggiolaro, Marco Antonio and Semini, Claudio},
  booktitle={2024 IEEE-RAS 23rd International Conference on Humanoid Robots (Humanoids)},
  pages={213--220},
  year={2024},
  organization={IEEE}
}

@inproceedings{kim2025adaptive,
  title={Adaptive Invariant Extended {Kalman} Filter for Legged Robot State Estimation},
  author={Kim, Kyung-Hwan and Ahn, DongHyun and Lee, Dong-hyun and Yoon, JuYoung and Hyun, Dong Jin},
  booktitle={2025 IEEE/RSJ International Conference on Intelligent Robots and Systems (IROS)},
  pages={3063--3068},
  year={2025},
  organization={IEEE}
}

@article{kim2021legged,
  title={Legged robot state estimation with dynamic contact event information},
  author={Kim, Joon-Ha and Hong, Seungwoo and Ji, Gwanghyeon and Jeon, Seunghun and Hwangbo, Jemin and Oh, Jun-Ho and Park, Hae-Won},
  journal={IEEE Robotics and Automation Letters},
  volume={6},
  number={4},
  pages={6733--6740},
  year={2021},
  publisher={IEEE}
}

@article{yoon2023invariant,
  title={Invariant smoother for legged robot state estimation with dynamic contact event information},
  author={Yoon, Ziwon and Kim, Joon-Ha and Park, Hae-Won},
  journal={IEEE Transactions on Robotics},
  volume={40},
  pages={193--212},
  year={2023},
  publisher={IEEE}
}

@article{nistico2025muse,
  title={{MUSE}: A Real-Time Multi-Sensor State Estimator for Quadruped Robots},
  author={Nistic{\`o}, Ylenia and Soares, Jo{\~a}o Carlos Virgolino and Amatucci, Lorenzo and Fink, Geoff and Semini, Claudio},
  journal={IEEE Robotics and Automation Letters},
  volume={10},
  number={5},
  pages={4620--4627},
  year={2025},
  publisher={IEEE}
}

@misc{girgin2026ocelotodometrycontactestimation,
      title={{OCELOT}: Odometry and Contact Estimation for Legged Robots}, 
      author={Emre Girgin and Cagri Kilic},
      year={2026},
      eprint={2605.21863},
      archivePrefix={arXiv},
      primaryClass={cs.RO},
      url={https://arxiv.org/abs/2605.21863}, 
}

@inproceedings{hohmeyer2025inekformer,
  title={{InEKFormer}: A Hybrid State Estimator for Humanoid Robots},
  author={Hohmeyer, Lasse and Popescu, Mihaela and Bergonzani, Ivan and Mronga, Dennis and Kirchner, Frank},
  booktitle={2025 IEEE International Conference on Advanced Robotics (ICAR)},
  pages={833--840},
  year={2025},
  organization={IEEE}
}

@inproceedings{fallon2014drift,
  title={Drift-free humanoid state estimation fusing kinematic, inertial and lidar sensing},
  author={Fallon, Maurice F and Antone, Matthew and Roy, Nicholas and Teller, Seth},
  booktitle={2014 IEEE-RAS International Conference on Humanoid Robots},
  pages={112--119},
  year={2014},
  organization={IEEE}
}

@inproceedings{rotella2018unsupervised,
  title={Unsupervised contact learning for humanoid estimation and control},
  author={Rotella, Nicholas and Schaal, Stefan and Righetti, Ludovic},
  booktitle={2018 IEEE International Conference on Robotics and Automation (ICRA)},
  pages={411--417},
  year={2018},
  organization={IEEE}
}

@article{camurri2017probabilistic,
  title={Probabilistic contact estimation and impact detection for state estimation of quadruped robots},
  author={Camurri, Marco and Fallon, Maurice and Bazeille, St{\'e}phane and Radulescu, Andreea and Barasuol, Victor and Caldwell, Darwin G and Semini, Claudio},
  journal={IEEE Robotics and Automation Letters},
  volume={2},
  number={2},
  pages={1023--1030},
  year={2017},
  publisher={IEEE}
}

@inproceedings{hwangbo2016probabilistic,
  title={Probabilistic foot contact estimation by fusing information from dynamics and differential/forward kinematics},
  author={Hwangbo, Jemin and Bellicoso, Carmine Dario and Fankhauser, P{\'e}ter and Hutter, Marco},
  booktitle={2016 IEEE/RSJ International Conference on Intelligent Robots and Systems (IROS)},
  pages={3872--3878},
  year={2016},
  organization={IEEE}
}

@inproceedings{bledt2018contact,
  title={Contact model fusion for event-based locomotion in unstructured terrains},
  author={Bledt, Gerardo and Wensing, Patrick M and Ingersoll, Sam and Kim, Sangbae},
  booktitle={2018 IEEE International Conference on Robotics and Automation (ICRA)},
  pages={4399--4406},
  year={2018},
  organization={IEEE}
}

@article{jenelten2019dynamic,
  title={Dynamic locomotion on slippery ground},
  author={Jenelten, Fabian and Hwangbo, Jemin and Tresoldi, Fabian and Bellicoso, C Dario and Hutter, Marco},
  journal={IEEE Robotics and Automation Letters},
  volume={4},
  number={4},
  pages={4170--4176},
  year={2019},
  publisher={IEEE}
}

@article{wu2025doglegs,
  title={{DogLegs}: Robust Proprioceptive State Estimation for Legged Robots Using Multiple Leg-Mounted {IMUs}},
  author={Wu, Yibin and Kuang, Jian and Khorshidi, Shahram and Niu, Xiaoji and Klingbeil, Lasse and Bennewitz, Maren and Kuhlmann, Heiner},
  journal={arXiv preprint arXiv:2503.04580},
  year={2025}
}

@article{baumgartner2026coco,
  title={{CoCo-InEKF}: State Estimation with Learned Contact Covariances in Dynamic, Contact-Rich Scenarios},
  author={Baumgartner, Michael and M{\"u}ller, David and Serifi, Agon and Grandia, Ruben and Knoop, Espen and Gross, Markus and B{\"a}cher, Moritz},
  journal={arXiv preprint arXiv:2605.15122},
  year={2026}
}

@article{ji2022concurrent,
  title={Concurrent training of a control policy and a state estimator for dynamic and robust legged locomotion},
  author={Ji, Gwanghyeon and Mun, Juhyeok and Kim, Hyeongjun and Hwangbo, Jemin},
  journal={IEEE Robotics and Automation Letters},
  volume={7},
  number={2},
  pages={4630--4637},
  year={2022},
  publisher={IEEE}
}

@inproceedings{yu2024state,
  title={State estimation transformers for agile legged locomotion},
  author={Yu, Chen and Yang, Yichu and Liu, Tianlin and You, Yangwei and Zhou, Mingliang and Xiang, Diyun},
  booktitle={2024 IEEE/RSJ International Conference on Intelligent Robots and Systems (IROS)},
  pages={6810--6817},
  year={2024},
  organization={IEEE}
}

@article{radford2018improving,
  title={Improving language understanding by generative pre-training},
  author={Radford, Alec and Narasimhan, Karthik and Salimans, Tim and Sutskever, Ilya and others},
  year={2018},
  publisher={San Francisco, CA, USA}
}

@article{lin2021legged,
  title={Legged robot state estimation using invariant {Kalman} filtering and learned contact events},
  author={Lin, Tzu-Yuan and Zhang, Ray and Yu, Justin and Ghaffari, Maani},
  journal={arXiv preprint arXiv:2106.15713},
  year={2021}
}

@article{sun2025proprioceptive,
  title={Proprioceptive slip detection and state estimation of multi-legged robots in slippery scenarios},
  author={Sun, Peng and Li, Qi and Hu, Hao and Qiang, Junjie and Wu, Weiwei and Luo, Xin},
  journal={Frontiers of Mechanical Engineering},
  volume={20},
  number={5},
  pages={36},
  year={2025},
  publisher={Springer}
}

@inproceedings{youm2025legged,
  title={Legged robot state estimation with invariant extended {Kalman} filter using neural measurement network},
  author={Youm, Donghoon and Oh, Hyunsik and Choi, Suyoung and Kim, Hyeongjun and Jeon, Seunghun and Hwangbo, Jemin},
  booktitle={2025 IEEE International Conference on Robotics and Automation (ICRA)},
  pages={670--676},
  year={2025},
  organization={IEEE}
}

@article{sola2017quaternion,
  title={Quaternion kinematics for the error-state {Kalman} filter},
  author={Sola, Joan},
  journal={arXiv preprint arXiv:1711.02508},
  year={2017}
}

@article{sola2018micro,
  title={A micro {Lie} theory for state estimation in robotics},
  author={Sola, Joan and Deray, Jeremie and Atchuthan, Dinesh},
  journal={arXiv preprint arXiv:1812.01537},
  year={2018}
}

@article{zimek2012survey,
  title={A survey on unsupervised outlier detection in high-dimensional numerical data},
  author={Zimek, Arthur and Schubert, Erich and Kriegel, Hans-Peter},
  journal={Statistical Analysis and Data Mining: The ASA Data Science Journal},
  volume={5},
  number={5},
  pages={363--387},
  year={2012},
  publisher={Wiley Online Library}
}

@inproceedings{jin2019joint,
  title={Joint torque estimation toward dynamic and compliant control for gear-driven torque sensorless quadruped robot},
  author={Jin, Bingchen and Sun, Caiming and Zhang, Aidong and Ding, Ning and Lin, Jing and Deng, Ganyu and Zhu, Zuwen and Sun, Zhenglong},
  booktitle={2019 IEEE/RSJ International Conference on Intelligent Robots and Systems (IROS)},
  pages={4630--4637},
  year={2019},
  organization={IEEE}
}

@inproceedings{patel2025tartanground,
  title={{TartanGround}: A large-scale dataset for ground robot perception and navigation},
  author={Patel, Manthan and Yang, Fan and Qiu, Yuheng and Cadena, Cesar and Scherer, Sebastian and Hutter, Marco and Wang, Wenshan},
  booktitle={2025 IEEE/RSJ International Conference on Intelligent Robots and Systems (IROS)},
  pages={20524--20531},
  year={2025},
  organization={IEEE}
}

@inproceedings{hutter2016anymal,
  title={{ANYmal} -- a highly mobile and dynamic quadrupedal robot},
  author={Hutter, Marco and Gehring, Christian and Jud, Dominic and Lauber, Andreas and Bellicoso, C Dario and Tsounis, Vassilios and Hwangbo, Jemin and Bodie, Karen and Fankhauser, Peter and Bloesch, Michael and others},
  booktitle={2016 IEEE/RSJ International Conference on Intelligent Robots and Systems (IROS)},
  pages={38--44},
  year={2016},
  organization={IEEE}
}

@article{mcinnes2018umap,
  title={{UMAP}: Uniform manifold approximation and projection for dimension reduction},
  author={McInnes, Leland and Healy, John and Melville, James},
  journal={arXiv preprint arXiv:1802.03426},
  year={2018}
}

@inproceedings{sturm2012benchmark,
  title={A benchmark for the evaluation of {RGB-D} {SLAM} systems},
  author={Sturm, J{\"u}rgen and Engelhard, Nikolas and Endres, Felix and Burgard, Wolfram and Cremers, Daniel},
  booktitle={2012 IEEE/RSJ International Conference on Intelligent Robots and Systems},
  pages={573--580},
  year={2012},
  organization={IEEE}
}

\clearpage
\appendix

\section{Error-State Extended Kalman Filter Design}
\label{appx:esekf}

Instead of directly filtering nominal states, we utilize ESEKF, which decouples the system into a non-linear nominal state $\mathbf{x}$ and a linear error state $\delta\mathbf{x}$. This formulation minimizes linearization errors because error states evolve more linearly than nominal states, particularly for variables defined on non-linear Lie group manifolds, such as orientation.

\noindent \textbf{Kinematics:} Forward kinematics $\mathbf{fk}_i$ and the Jacobian $\mathbf{J}_{v,i}$ transform the joint configuration $\mathbf{q}_i \in \mathbb{R}^N$ and joint velocities $\dot{\mathbf{q}}_i$ of leg $i$ into the foot position $\mathbf{p}_i^B \in \mathbb{R}^3$ and linear velocity $\mathbf{v}_i^B \in \mathbb{R}^3$ relative to the body frame $\{B\}$:
\begin{equation}
    \mathbf{p}_i^B = \mathbf{fk}_i(\mathbf{q}_i), \quad \mathbf{v}_i^B = \mathbf{J}_{v,i}(\mathbf{q}_i) \dot{\mathbf{q}}_i
\end{equation}

\noindent \textbf{State Definition:} The nominal state $\mathbf{x}$ tracks world-frame position $\mathbf{p}^W \in \mathbb{R}^3$ and velocity $\mathbf{v}^W \in \mathbb{R}^3$, the $\{B\}$ to $\{W\}$ rotation matrix $\mathbf{R} \in SO(3)$, alongside additive body-frame accelerometer and gyroscope biases $\mathbf{b}_a, \mathbf{b}_g \in \mathbb{R}^3$. The corresponding error state $\delta\mathbf{x}$ is modeled as a zero-mean Gaussian:
\begin{equation}
    \delta\mathbf{x} = [\delta\mathbf{p}^T, \delta\mathbf{v}^T, \delta\boldsymbol{\theta}^T, \delta\mathbf{b}_a^T, \delta\mathbf{b}_g^T]^T \in \mathbb{R}^{15}
\end{equation}
True states are synthesized by injecting the error state into the nominal state, utilizing the exponential map $\text{Exp}(\cdot): \mathbb{R}^3 \to SO(3)$ for rotation matrices \citep{sola2018micro}:
\begin{equation}
    \mathbf{p}_{\text{true}} = \mathbf{p}^W + \delta\mathbf{p}, \quad
    \mathbf{v}_{\text{true}} = \mathbf{v}^W + \delta\mathbf{v}, \quad
    \mathbf{R}_{\text{true}} = \mathbf{R}\text{Exp}(\delta\boldsymbol{\theta}), \quad
    \mathbf{b}_{a, \text{true}} = \mathbf{b}_a + \delta\mathbf{b}_a, \quad
    \mathbf{b}_{g, \text{true}} = \mathbf{b}_g + \delta\mathbf{b}_g
\end{equation}

\noindent \textbf{Prediction Step (Propagation):}
Nominal states advance by integrating raw IMU readings, $\mathbf{a}_{\text{raw}}$ and $\boldsymbol{\omega}_{\text{raw}}$, across the time step $\Delta t$. Bias compensation and Euler integration proceed simultaneously, incorporating the global gravity vector $\mathbf{g}^W = [0, 0, -g]^T$:
\begin{equation}
\begin{aligned}
    \boldsymbol{\omega}^B &= \boldsymbol{\omega}_{\text{raw}} - \mathbf{b}_{g, k-1} &\qquad  \mathbf{R}_k &= \mathbf{R}_{k-1} \cdot \text{Exp}(\boldsymbol{\omega}^B \Delta t) \\
    \mathbf{a}^B &= \mathbf{a}_{\text{raw}} - \mathbf{b}_{a, k-1} &\qquad  \mathbf{v}_k^W &= \mathbf{v}_{k-1}^W + \mathbf{a}^W \Delta t \\
    \mathbf{a}^W &= \mathbf{R}_{k-1} \mathbf{a}^B + \mathbf{g}^W &\qquad  \mathbf{p}_k^W &= \mathbf{p}_{k-1}^W + \mathbf{v}_{k-1}^W \Delta t + 0.5 \mathbf{a}^W \Delta t^2
\end{aligned}
\end{equation}

Error covariance propagation follows $\mathbf{P}_k = \mathbf{F}_d \mathbf{P}_{k-1} \mathbf{F}_d^T + \mathbf{Q}_d$. The discrete transition matrix $\mathbf{F}_d \approx \mathbf{I} + \mathbf{F}_c \Delta t$ and discrete process noise $\mathbf{Q}_d = \mathbf{G} \mathbf{Q}_c \mathbf{G}^T \Delta t$ originate from the continuous-time error dynamics $\mathbf{F}_c$ and noise mapping $\mathbf{G}$ ($[\cdot]_\times$ denotes the skew-symmetric operator, and $\mathbf{Q}_c \in \mathbb{R}^{12 \times 12}$ is the continuous noise covariance):
\begin{equation}
    \setlength{\arraycolsep}{2.5pt} 
    \mathbf{F}_c =
    \begin{bmatrix}
        \mathbf{0} & \mathbf{I} & \mathbf{0} & \mathbf{0} & \mathbf{0} \\
        \mathbf{0} & \mathbf{0} & - \mathbf{R} [\mathbf{a}^B]_\times & - \mathbf{R} & \mathbf{0} \\
        \mathbf{0} & \mathbf{0} & -[\boldsymbol{\omega}^B]_\times & \mathbf{0} & -\mathbf{I} \\
        \mathbf{0} & \mathbf{0} & \mathbf{0} & \mathbf{0} & \mathbf{0} \\
        \mathbf{0} & \mathbf{0} & \mathbf{0} & \mathbf{0} & \mathbf{0}
    \end{bmatrix}, \quad
    \mathbf{G} =
    \begin{bmatrix}
        \mathbf{0} & \mathbf{0} & \mathbf{0} & \mathbf{0} \\
        -\mathbf{R} & \mathbf{0} & \mathbf{0} & \mathbf{0} \\
        \mathbf{0} & -\mathbf{I} & \mathbf{0} & \mathbf{0} \\
        \mathbf{0} & \mathbf{0} & \mathbf{I} & \mathbf{0} \\
        \mathbf{0} & \mathbf{0} & \mathbf{0} & \mathbf{I}
    \end{bmatrix}
\end{equation}

\noindent \textbf{Correction Step (Zero-Velocity Update):}
The fundamental kinematic assumption of leg odometry is that the stance foot remains static within the world frame $\{W\}$, establishing the constraint $\mathbf{v}_{\text{foot}, i}^W = \mathbf{0}$. The measurement function $h(\mathbf{x}_k)$ computes the predicted world-frame velocity of the stance foot by superimposing the base velocity $\mathbf{v}_k^W$ with the body-relative foot velocity $\mathbf{v}_{\text{rel},i}^B$. For each foot $i$, measurement Jacobian $\mathbf{H}_k = \partial h / \partial \delta\mathbf{x} \in \mathbb{R}^{3 \times 15}$ is formulated as:
\begin{equation} \label{eqn:measurement_model}
\begin{aligned}
    h(\mathbf{x}_k) &= \mathbf{v}_k^W + \mathbf{R}_k \mathbf{v}_{\text{rel},i}^B, \quad \mathbf{v}_{\text{rel},i}^B = \boldsymbol{\omega}^B \times \mathbf{p}_i^B + \mathbf{J}_{v,i}(\mathbf{q}_i)\dot{\mathbf{q}}_i \\
    \mathbf{H}_k &= \begin{bmatrix} \mathbf{0}_{3\times3} & \mathbf{I}_{3\times3} & - \mathbf{R}_k [\mathbf{v}_{\text{rel},i}^B]_\times & \mathbf{0}_{3\times3} & \mathbf{R}_k [\mathbf{p}_i^B]_\times \end{bmatrix}
\end{aligned}
\end{equation}

Given the pseudo-measurement $\mathbf{y}_k = \mathbf{0}$, the resulting innovation $\mathbf{\nu}_k = -h(\mathbf{x}_k)$ undergoes a Mahalanobis distance check to filter out non-stationary events like slippage \citep{6697236} or stance with high covariance. Updates are aborted if $\mathbf{\nu}_k^T \mathbf{S}_k^{-1} \mathbf{\nu}_k > \gamma_{95}$, where $\mathbf{S}_k = \mathbf{H}_k \mathbf{P}_k \mathbf{H}_k^T + \mathbf{R}_k$ defines the innovation covariance. Upon acceptance, the optimal Kalman gain $\mathbf{K}_k$ is evaluated to extract the state correction $\delta\mathbf{x}^+$, and the posterior covariance $\mathbf{P}_k^+$ is updated via the numerically stable Joseph form:
\begin{equation}
\begin{aligned}
    \mathbf{K}_k &= \mathbf{P}_k \mathbf{H}_k^T \mathbf{S}_k^{-1}, \quad \delta\mathbf{x}^+ = \mathbf{K}_k \mathbf{\nu}_k \\
    \mathbf{P}_k^+ &= (\mathbf{I} - \mathbf{K}_k \mathbf{H}_k) \mathbf{P}_k (\mathbf{I} - \mathbf{K}_k \mathbf{H}_k)^T + \mathbf{K}_k \mathbf{R}_{k} \mathbf{K}_k^T
\end{aligned}
\end{equation}
Ultimately, the nominal state absorbs the calculated corrections. Linear variables are updated via standard addition, whereas the rotation matrix relies on the Lie group exponential map:
\begin{equation}
    \mathbf{R}^+ = \mathbf{R} \text{Exp}(\delta\boldsymbol{\theta}^+) , \quad \mathbf{z}^+ = \mathbf{z} + \delta\mathbf{z}^+ \;\; \forall \mathbf{z} \in \{\mathbf{p}, \mathbf{v}, \mathbf{b}_a, \mathbf{b}_g\}
\end{equation}

\section{Kinematic Feature Selection}
\label{sec:feature_selection}

\begin{figure}[h]
    \centering
    \includegraphics[width=1\linewidth]{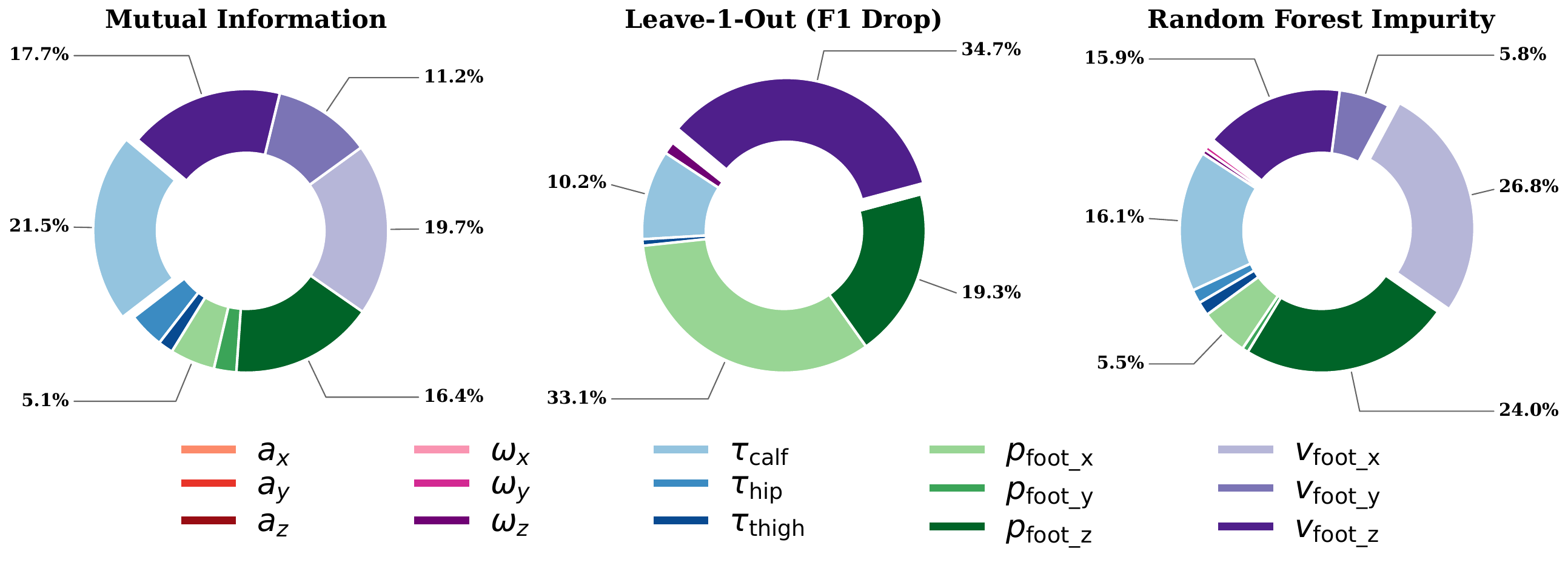}
    \caption{Feature Selection outputs ranking the predictive power of kinematic variables relative to the binary Ground Reaction Force (GRF) contact state.}
    \label{fig:feature_selection}
\end{figure}

The standard proprioceptive sensor suite for quadrupedal robots (joint encoders and IMU) yields a dense observation space comprising joint angles ($\mathbf{q}$), velocities ($\mathbf{\dot{q}}$), accelerations ($\mathbf{\ddot{q}}$), estimated torques ($\boldsymbol{\tau}$) \citep{jin2019joint}, base linear acceleration ($\mathbf{a}$), and angular velocity ($\boldsymbol{\omega}$). To isolate a minimal, high-signal kinematic subset $\mathcal{S}$ capable of reproducing the GRF-derived contact distributions, we executed a tripartite feature selection protocol directly against binary GRF pseudo-labels:

\noindent \textbf{Mutual Information (MI):}  A non-parametric filter method quantifying the non-linear dependency (in bits) between individual continuous kinematic signals and the discrete contact labels.

\noindent \textbf{Random Forest (RF) Importance:} A predictive ensemble method ranking features based on the Mean Decrease in Gini Impurity when trained to classify the contact state.

\noindent \textbf{Leave-One-Out (L1O) Ablation:} A wrapper method evaluating the structural importance of features within the generative model. The baseline 2-state HMM-GMM is iteratively re-fitted with one feature masked. The structural importance is quantified by the absolute degradation in the F1-score relative to the full-state baseline.

As illustrated in Figure \ref{fig:feature_selection}, higher-order derivatives susceptible to high-frequency noise exhibit negligible predictive utility. The consensus across all three metrics identified an optimal, compact 5-dimensional subset: vertical foot position relative to the base ($p_z^B$), 3D foot velocity ($\mathbf{v}_{\text{rel}}^B$), and estimated calf torque ($\tau_{\text{calf}}$). This subset serves as the observation vector $o_t$ for all subsequent estimation frameworks. Note that the static-stance standard deviation of the estimated calf torque lies within $\sigma_{\tau} \in [0.0480, 0.1169]\,\text{N}\cdot\text{m}$.

\section{Contact Detectors}
\subsection{HMM-GMM}
\label{appx:hmm-gmm}

State estimation is performed online via a recursive Bayesian filter. The belief $p_t(j) = P(x_t=j \mid o_{1:t})$ is updated at each time step $t$ using the forward algorithm:
\begin{equation}
p_t(j)=\eta \, b_j(o_t)\sum_{i\in\mathcal{X}}p_{t-1}(i)a_{ij},
\end{equation}
where $\eta$ is a normalization factor ensuring $\sum_{j\in\mathcal{X}} p_t(j) = 1$, and $a_{ij} = P(x_t=j\mid x_{t-1}=i)$. The transition probabilities $A$ are held constant and act as a tuning parameter to regulate the filter's sensitivity to state switching.

We have two operational modes for fitting the emission probabilities to address deployment constraints. The \textit{Offline} (Batch) mode fits GMM once over the entire trajectory prior to state estimation. This provides a global, robust, sequence specific emission model but lacks real-time adaptability. On the other hand, \textit{Online} (Sliding Window) approach is utilized for real-time deployment. The GMM is continuously re-fitted over a sliding window of length $N$ at a specified update frequency. The sliding window enables real-time performance and allows the emission probabilities to dynamically adapt to varying terrain properties. Degenerate cases violating the bimodal distribution assumption, such as staying stationary, are detected when the largest mixture weight exceeds 0.98, in which case the system reverts to a precomputed nominal GMM. The transition matrix is $A = \begin{bmatrix} 0.99 & 0.01 \\ 0.01 & 0.99 \end{bmatrix}$ and the initial belief is uniform, $\pi = [0.5, 0.5]$.

\subsection{Supervised Architectures}
\label{appx:supervised}

Our CNN based architecture is a 1D CNN that processes the sequence using two consecutive convolutional blocks. We follow the exact architecture of \citet{lin2021legged} for comparison. Each block consists of two stacked convolutional layers (kernel size 3, padding configured to preserve length) followed by max pooling (kernel size 2, stride 2) and dropout ($p=0.5$). The pooled temporal features are flattened and passed through a Multi-Layer Perceptron (MLP) with hidden dimensions $[2048, 512, 1]$.

Our other architecture is a recurrent network that processes the input sequentially using a single-layer GRU with a hidden state dimension of 128. To isolate the current state prediction, only the hidden state of the final sequence timestep is extracted. This vector is processed by an MLP with hidden dimensions $[256, 128, 1]$ utilizing dropout ($p=0.5$) between linear layers. As with the CNN, we follow the exact architecture of \citet{youm2025legged}. 

\subsection{Self-Supervised Architectures}
\label{appx:self-supervised}

The autoencoder processes a kinematic input tensor $\mathbf{X} \in \mathbb{R}^{B \times W \times C}$ (batch size $B$, window size $W$, feature dimension $C$) through an asymmetric sequence-to-sequence topology, utilizing one of two mutually exclusive encoders to generate a latent vector $\mathbf{z} \in \mathbb{R}^{16}$.

\noindent \textbf{Encoders:} The CNN encoder is a 3-layer 1D convolutional stack with $C_{out}=\{32, 64, 128\}$ channels, kernel sizes $K=\{5, 5, 3\}$, and strides $S=\{2, 2, 2\}$, each followed by a GELU activation, and the final feature map is flattened and linearly projected to $\mathbf{z}$. The GRU encoder is a single-layer GRU with hidden dimension 64, whose terminal hidden state $\mathbf{h}_{W}$ is linearly projected to $\mathbf{z}$.

\noindent \textbf{Decoder:} The latent vector $\mathbf{z}$ is linearly expanded and reshaped into $\mathbf{H}_{0} \in \mathbb{R}^{128 \times L_{base}}$ and upsampled to the original dimensions by three 1D transposed convolutions with channels $128 \rightarrow 64 \rightarrow 32 \rightarrow 5$, kernel sizes $\{3, 5, 5\}$, and stride 2. The first two layers use GELU activations and the last one is linear. To enforce exact dimensional alignment, the base sequence length $L_{base}$ and the output paddings of the transposed convolutions are determined via exhaustive search during network initialization.

\section{Experimental Details}
\label{appx:experiments}

In our experiments, the train/val/test sets are disjoint, with ratios of 70/15/15\%, respectively. All models are optimized using Adam optimizer on train set and best checkpoint is decided using lowest validation loss, per dataset. Hyperparameters are selected via the best performing configuration at the downstream leg odometry task on validation set. The models are trained per dataset and the sim-to-real gap is not tested.

\noindent \textbf{TartanGround (Simulation):} 240 sequences from the TartanGround dataset \citep{patel2025tartanground} collected using the ANYmal robot \citep{hutter2016anymal}. This comprises 8 hours and 23.5 km of locomotion data across 6 terrain profiles. Ground truth contact states are derived from noise-free simulated GRF ($F_z$). 

\noindent \textbf{Real-World Domain:} Comprises 25 minutes and 1.10 km of data recorded on the Unitree Go2 quadruped traversing concrete (470\,m), grass (500\,m), and rock (130\,m), which is subject to slippage frequently. Two distinct locomotion controllers are used to induce gait diversity. Absolute time synchronized RTK GPS (5 Hz, sub-centimeter precision) provides ground truth trajectories for odometry evaluation. For the real-world dataset, pseudo-labels are generated via an offline HMM-GMM method fitted exclusively to the 1D GRF signal, as standard hard thresholding fails under sensor noise during the swing phase. The dataset is available in the project repository.

\noindent \textbf{Default Hyperparameters:} By default, we set window size ($W$) 20 for supervised CNN and GRU and 1 for DAE-GMM (CNN) and DAE-GMM (GRU), effectively reducing the CNN to an MLP and the GRU to a single recurrent step. However, we keep the architecture the same to experiment over varying window sizes in our ablation experiments. The latent dimension is 16 for both DAE-GMM (CNN) and DAE-GMM (GRU). The GMMs of the DAE-GMMs are fitted during DAE training and kept fixed during inference. Therefore, both DAE-GMMs are online by design. HMM-GMM method's online setting used sliding window of size 500 and refitted at every 250 samples. GRF thresholds are 3 N and 30 N for simulation and real-world datasets, respectively. For GRF thresholding, $p_t$ is calculated as $F_z/\mathbb{T}$, clamped between 0 and 1. 

\noindent \textbf{Implementation Details:} The foot velocity feature is computed as $\mathbf{v}_{\text{rel}}^B = \boldsymbol{\omega}\times\mathbf{p}^B + \mathbf{J}_v(\mathbf{q})\dot{\mathbf{q}}$, using the gyroscope and the joint velocities reported by the motor drivers (no numerical differentiation in our pipeline), and the calf torque is the driver's current-based estimate. The Jacobian is analytical for ANYmal and computed by central differences of the forward kinematics for Go2. HMM-GMM uses full-covariance GMMs (200 EM iterations, $10^{-6}$ ridge) on unstandardized features. In offline mode, one GMM is fitted per leg, and in online mode a window is treated as degenerate, reverting to a GMM pre-fitted on the training set, when the largest mixture weight exceeds 0.98. The DAE inputs are standardized with training-set statistics and corrupted with Gaussian noise ($\sigma=0.01$) and per-sample amplitude scaling $\sim\mathcal{U}[0.98, 1.02]$ before being passed to the network, which is trained with Adam (learning rate $10^{-3}$, batch size 1024, 30 epochs), keeping the checkpoint with the lowest validation MSE. A single full-covariance GMM, shared across legs, is fitted to the training latents. The ESEKF uses $\sigma_a = 0.5$\,m/s$^2$, $\sigma_g = 1^\circ$/s, $\sigma_{b_a} = 0.01$, $\sigma_{b_g} = 0.01^\circ$/s. NIS gating is applied per foot, and all accepted feet are fused in one stacked update.

\subsection{Evaluation Metric Definitions}
\label{sec:metrics}

Let the ground truth trajectory be defined as a sequence of poses $P = \{p_1, p_2, \dots, p_N\}$, where $p_i \in \mathbb{R}^d$ ($d \in \{2, 3\}$). Let the estimated trajectory, synchronized to the ground truth timestamps via linear interpolation, be $\hat{P} = \{\hat{p}_1, \hat{p}_2, \dots, \hat{p}_N\}$. The initial states are perfectly aligned such that $p_1 = \hat{p}_1$ and the initial coordinate frames are identical.

For angular metrics, let $\theta_i$ and $\hat{\theta}_i$ denote the headings derived from the sequential spatial displacements of $P$ and $\hat{P}$, respectively. Let $\text{wrap}(\cdot)$ denote angular normalization to the interval $[-\pi, \pi]$.

\subsubsection{Absolute Trajectory Error (ATE)}
ATE evaluates the global consistency of the trajectory by computing the Root Mean Square Error (RMSE) of the synchronized Euclidean position differences.
\begin{equation}
    \text{ATE} = \sqrt{\frac{1}{N} \sum_{i=1}^{N} \lVert p_i - \hat{p}_i \rVert_2^2}
\end{equation}

\subsubsection{Absolute Heading Error (AHE)}
AHE computes the RMSE of the absolute difference between the ground truth and predicted headings.
\begin{equation}
    \text{AHE} = \sqrt{ \frac{1}{N-1} \sum_{i=1}^{N-1} \text{wrap}(\hat{\theta}_i - \theta_i)^2 }
\end{equation}

\subsubsection{Relative Pose Error (RPE)}
RPE measures the local drift over a predefined interval. Let $S$ denote the set of valid index pairs $(i, j)$ such that the traveled path distance between $p_i$ and $p_j$ corresponds to a fixed evaluation window $\Delta$. Let $R(\theta) \in SO(d)$ represent the rotation matrix constructed from heading $\theta$.

The translational RPE evaluates the relative spatial drift by comparing local displacements, normalized by the interval distance, expressed as a percentage:
\begin{equation}
    \text{RPE}_{\text{trans}} = \sqrt{ \frac{1}{|S|} \sum_{(i,j) \in S} \left( \frac{\lVert R(\theta_i)^{-1}(p_j - p_i) - R(\hat{\theta}_i)^{-1}(\hat{p}_j - \hat{p}_i) \rVert_2}{\Delta} \right)^2 } \times 100
\end{equation}

The rotational RPE evaluates the local heading drift per unit distance, expressed in degrees per meter:
\begin{equation}
    \text{RPE}_{\text{rot}} = \sqrt{ \frac{1}{|S|} \sum_{(i,j) \in S} \left( \frac{\text{wrap}\big((\hat{\theta}_j - \hat{\theta}_i) - (\theta_j - \theta_i)\big)}{\Delta} \right)^2 } \times \frac{180}{\pi}
\end{equation}

\subsubsection{Final Position Error (FPE)}
FPE computes the terminal drift of the system.
\begin{equation}
    \text{FPE} = \lVert p_N - \hat{p}_N \rVert_2
\end{equation}

\subsubsection{Discrete Fréchet Distance}
The discrete Fréchet distance evaluates the structural shape similarity of the paths independent of temporal alignment. Let $A$ and $B$ represent the discrete spatial sequences of $P$ and $\hat{P}$. Let $C$ represent the set of all monotonically ordered index couplings $(u_k, v_k)$ that traverse $A$ and $B$.
\begin{equation}
    \text{Fréchet}(A, B) = \min_{C} \max_{(u_k, v_k) \in C} \lVert A_{u_k} - B_{v_k} \rVert_2
\end{equation}

\section{Additional Experiments}
\label{appx:additional}

This section first reports two ablations on the design of the DAE (Appendices \ref{appx:window_size} and \ref{appx:latent_dim}). It then analyzes the stance posterior and its use in the ESEKF: calibration and reliability (Appendix \ref{appx:calibration}), sensitivity to the covariance scaling (Appendix \ref{appx:covariance_sensitivity}), the number of GMM components (Appendix \ref{appx:num_components}), and the cluster-to-state assignment under large elevation changes (Appendix \ref{appx:cluster_assignment}).

\subsection{Effect of Window Size ($W$)}
\label{appx:window_size}

Consistent with our analysis of the latent space (Figure \ref{fig:window_size_analysis}), the best performance across all metrics is achieved with the smallest window size ($W=1$). Table \ref{tab:tartanground_window_size_comparison} summarizes the odometry errors for different window sizes.

\begin{table}[htbp]
\centering
\caption{Effect of Window Size $W$, DAE-GMM (CNN), TartanGround $\downarrow$ (\textbf{Best}, \underline{Second Best})}
\label{tab:tartanground_window_size_comparison}
\small
\begin{tabular}{lcccccc}
\toprule
\textbf{Window Size} & \textbf{ATE} & \textbf{AHE} & \textbf{RPE Trans} & \textbf{RPE Rot} & \textbf{FPE} & \textbf{Fréchet} \\
 & \textbf{(m)} & \textbf{($^\circ$)} & \textbf{(\%)} & \textbf{($^\circ$/m)} & \textbf{(m)} & \textbf{(m)} \\
\midrule
1   & \textbf{4.09} & \textbf{2.69} & \textbf{9.93} & \textbf{0.141} & \textbf{7.13} & \textbf{7.28} \\
5   & 19.83 & 22.38 & 40.60 & 0.581 & 32.69 & 33.37 \\
20  & \underline{12.15} & \underline{6.14} & \underline{23.96} & \underline{0.236} & \underline{18.73} & \underline{19.53} \\
50  & 43.95 & 55.35 & 80.08 & 1.319 & 66.76 & 67.72 \\
100 & 51.47 & 92.08 & 98.70 & 3.153 & 85.24 & 87.81 \\
\bottomrule
\end{tabular}
\end{table}

\FloatBarrier
\subsection{Effect of Latent Dimensionality ($D$)}
\label{appx:latent_dim}

\begin{figure}[htbp]
    \centering
    \begin{subfigure}[b]{0.18\textwidth}
        \centering
        \includegraphics[width=\textwidth]{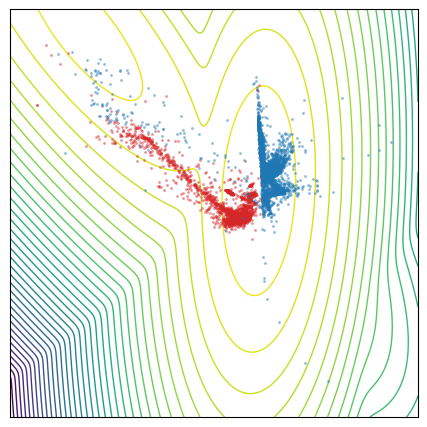}
    \end{subfigure}
    \hfill
    \begin{subfigure}[b]{0.18\textwidth}
        \centering
        \includegraphics[width=\textwidth, angle=0,  origin=c]{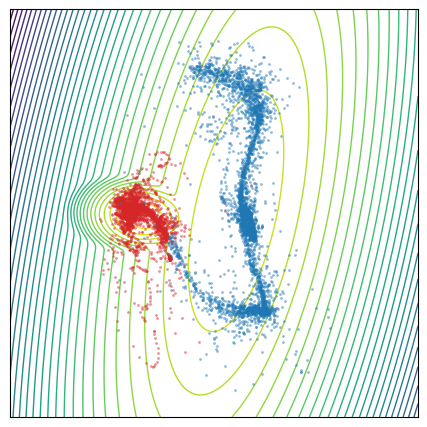}
    \end{subfigure}
    \hfill
    \begin{subfigure}[b]{0.18\textwidth}
        \centering
        \includegraphics[width=\textwidth, angle=180, origin=c]{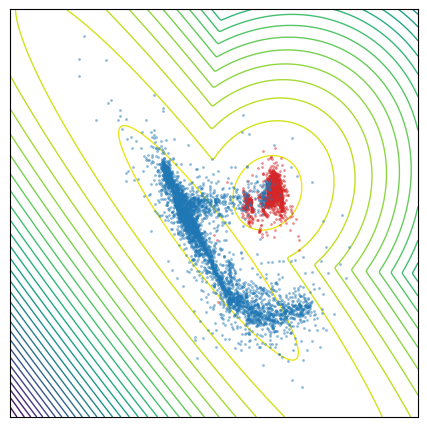}
    \end{subfigure}
    \hfill
    \begin{subfigure}[b]{0.18\textwidth}
        \centering
        \includegraphics[width=\textwidth]{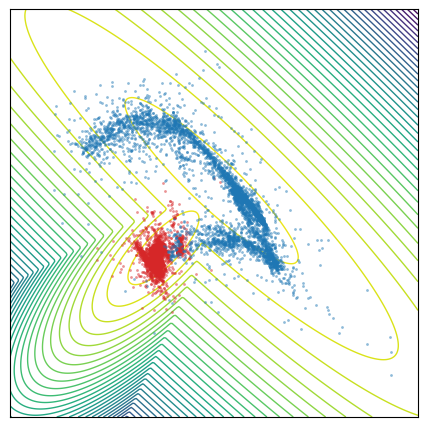}
    \end{subfigure}
    \hfill
    \begin{subfigure}[b]{0.18\textwidth}
        \centering
        \reflectbox{\includegraphics[width=\textwidth, angle=180, origin=c]{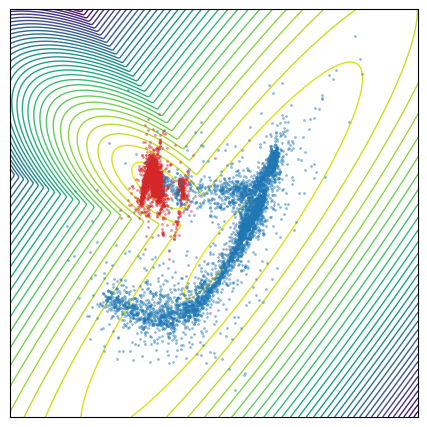}}
    \end{subfigure}

    \begin{subfigure}[b]{0.18\textwidth}
        \centering
        \includegraphics[width=\textwidth]{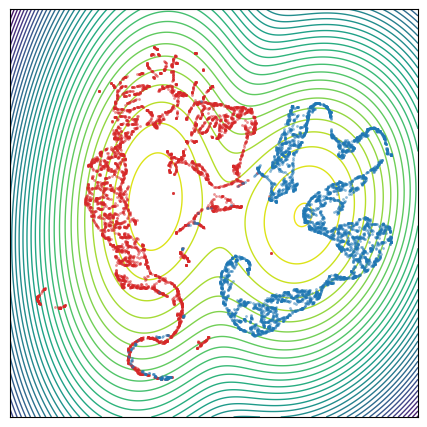}
        \caption{$D=2$}
    \end{subfigure}
    \hfill
    \begin{subfigure}[b]{0.18\textwidth}
        \centering
        \reflectbox{\includegraphics[width=\textwidth, angle=180,  origin=c]{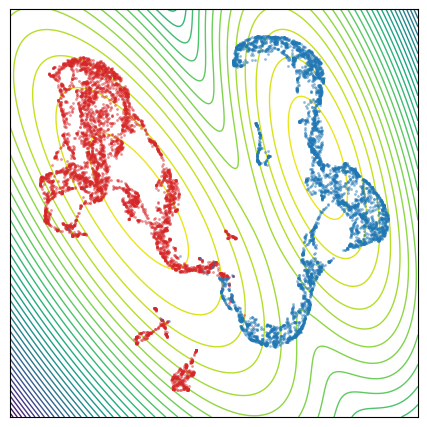}}
        \caption{$D=4$}
    \end{subfigure}
    \hfill
    \begin{subfigure}[b]{0.18\textwidth}
        \centering
        \reflectbox{\includegraphics[width=\textwidth]{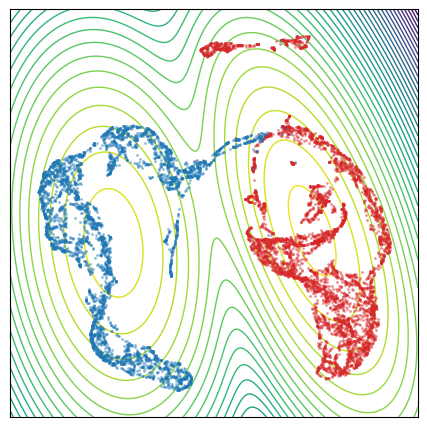}}
        \caption{$D=8$}
    \end{subfigure}
    \hfill
    \begin{subfigure}[b]{0.18\textwidth}
        \centering
        \reflectbox{\includegraphics[width=\textwidth, angle=180, origin=c]{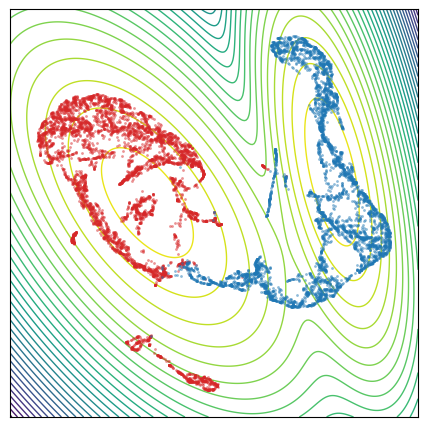}}
        \caption{$D=16$}
    \end{subfigure}
    \hfill
    \begin{subfigure}[b]{0.18\textwidth}
        \centering
        \reflectbox{\includegraphics[width=\textwidth, angle=180, origin=c]{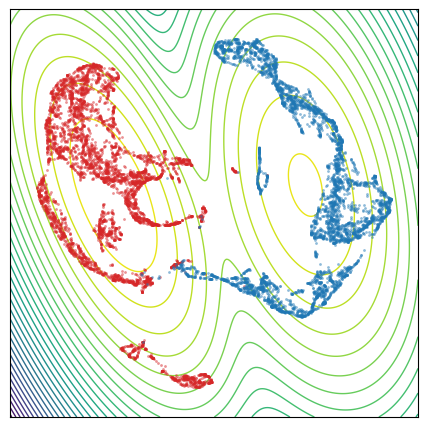}}
        \caption{$D=32$}
    \end{subfigure}

    \caption{Latent space for different latent dimensions $D$, projected onto PCA (top) and UMAP (bottom) dimensions. Colors indicate pseudo contact labels (red for stance, blue for swing).}
    \label{fig:latent_dim_analysis}
\end{figure}

We evaluated the DAE latent space across $D \in \{2,4,8,16,32,64\}$. Visually (Figure \ref{fig:latent_dim_analysis}), $D < 4$ forces excessive compression, while $D > 4$ does not provide additional separation in the latent space. Since $D=16$ yields the lowest errors for DAE-GMM (CNN), we keep $D=16$ to keep our hyperparameter search protocol consistent. Tables \ref{tab:ld_cnn} and \ref{tab:ld_gru} report the effect of the latent size on TartanGround. These runs use different random seeds than the main experiments (Section \ref{sec:experiments}), so the $D=16$ rows differ slightly from Table \ref{tab:tartanground_contact_detector_comparison}.

\begin{table}[htbp]
\centering
\caption{Effect of Latent Dimension $D$, DAE-GMM (CNN), TartanGround $\downarrow$ (\textbf{Best}, \underline{Second Best})}
\label{tab:ld_cnn}
\small
\begin{tabular}{lcccccc}
\toprule
\textbf{Latent Dim.} & \textbf{ATE} & \textbf{AHE} & \textbf{RPE Trans} & \textbf{RPE Rot} & \textbf{FPE} & \textbf{Fréchet} \\
 & \textbf{(m)} & \textbf{($^\circ$)} & \textbf{(\%)} & \textbf{($^\circ$/m)} & \textbf{(m)} & \textbf{(m)} \\
\midrule
2  & 13.99 & 17.03 & 31.98 & 0.462 & 24.13 & 24.70 \\
4  & 4.72  & 3.54  & 10.75 & 0.165 & 7.50  & 7.67  \\
8  & \underline{4.13} & 2.88 & 10.06 & 0.142 & \underline{7.24} & \underline{7.39} \\
16 & \textbf{4.13} & \textbf{2.79} & \textbf{10.00} & \underline{0.141} & \textbf{7.21} & \textbf{7.39} \\
32 & 4.16  & \underline{2.87} & \underline{10.02} & \textbf{0.141} & 7.28  & 7.46  \\
64 & 13.54 & 15.67 & 27.88 & 0.452 & 22.33 & 22.75 \\
\bottomrule
\end{tabular}

\vspace{1em}
\caption{Effect of Latent Dimension $D$, DAE-GMM (GRU), TartanGround $\downarrow$ (\textbf{Best}, \underline{Second Best})}
\label{tab:ld_gru}
\small
\begin{tabular}{lcccccc}
\toprule
\textbf{Latent Dim.} & \textbf{ATE} & \textbf{AHE} & \textbf{RPE Trans} & \textbf{RPE Rot} & \textbf{FPE} & \textbf{Fréchet} \\
 & \textbf{(m)} & \textbf{($^\circ$)} & \textbf{(\%)} & \textbf{($^\circ$/m)} & \textbf{(m)} & \textbf{(m)} \\
\midrule
2  & 5.33  & 3.82  & 11.99 & 0.166 & 8.60  & 8.87  \\
4  & 4.62  & \textbf{2.54} & 10.70 & \textbf{0.139} & 7.93  & 8.14  \\
8  & 4.16  & 2.82  & \underline{10.06} & 0.141 & 7.27  & 7.46  \\
16 & 4.19  & 2.90  & 10.12 & 0.142 & 7.31  & 7.50  \\
32 & \underline{4.15} & 2.85  & 10.06 & 0.142 & \underline{7.26} & \underline{7.46} \\
64 & \textbf{4.13} & \underline{2.78} & \textbf{9.99}  & \underline{0.141} & \textbf{7.22} & \textbf{7.39} \\
\bottomrule
\end{tabular}
\end{table}

\subsection{Calibration and Reliability of the Stance Posterior}
\label{appx:calibration}

\begin{figure}[htbp]
    \centering
    \begin{minipage}[c]{0.40\linewidth}
        \centering
        \footnotesize
        \setlength{\tabcolsep}{3pt}
        \renewcommand{\arraystretch}{0.9}
        \begin{tabular}{r r c c}
            \toprule
            \textbf{Bin} & \multicolumn{1}{c}{\textbf{$N$}} & \textbf{Mean $P$} & \textbf{GRF Freq.} \\
            \midrule
            $[0.0, 0.1]$ &  6,734 & 0.002 & 0.185 \\
            $[0.1, 0.2]$ &    138 & 0.147 & 0.739 \\
            $[0.2, 0.3]$ &     65 & 0.246 & 0.738 \\
            $[0.3, 0.4]$ &     44 & 0.340 & 0.727 \\
            $[0.4, 0.5]$ &     34 & 0.444 & 0.824 \\
            $[0.5, 0.6]$ &     48 & 0.556 & 0.813 \\
            $[0.6, 0.7]$ &     46 & 0.645 & 0.913 \\
            $[0.7, 0.8]$ &     61 & 0.751 & 0.853 \\
            $[0.8, 0.9]$ &     80 & 0.857 & 0.800 \\
            $[0.9, 1.0]$ & 12,378 & 0.999 & 0.985 \\
            \bottomrule
        \end{tabular}
        \captionof{table}{Reliability Table of the Stance Posterior Against GRF Pseudo-Labels. $N$: samples per bin. Mean $P$: mean predicted stance probability. GRF Freq.: fraction of samples labeled as stance.}
        \label{tab:binned_summary}
    \end{minipage}
    \hfill
    \begin{minipage}[c]{0.56\linewidth}
        \centering
        \includegraphics[width=\linewidth]{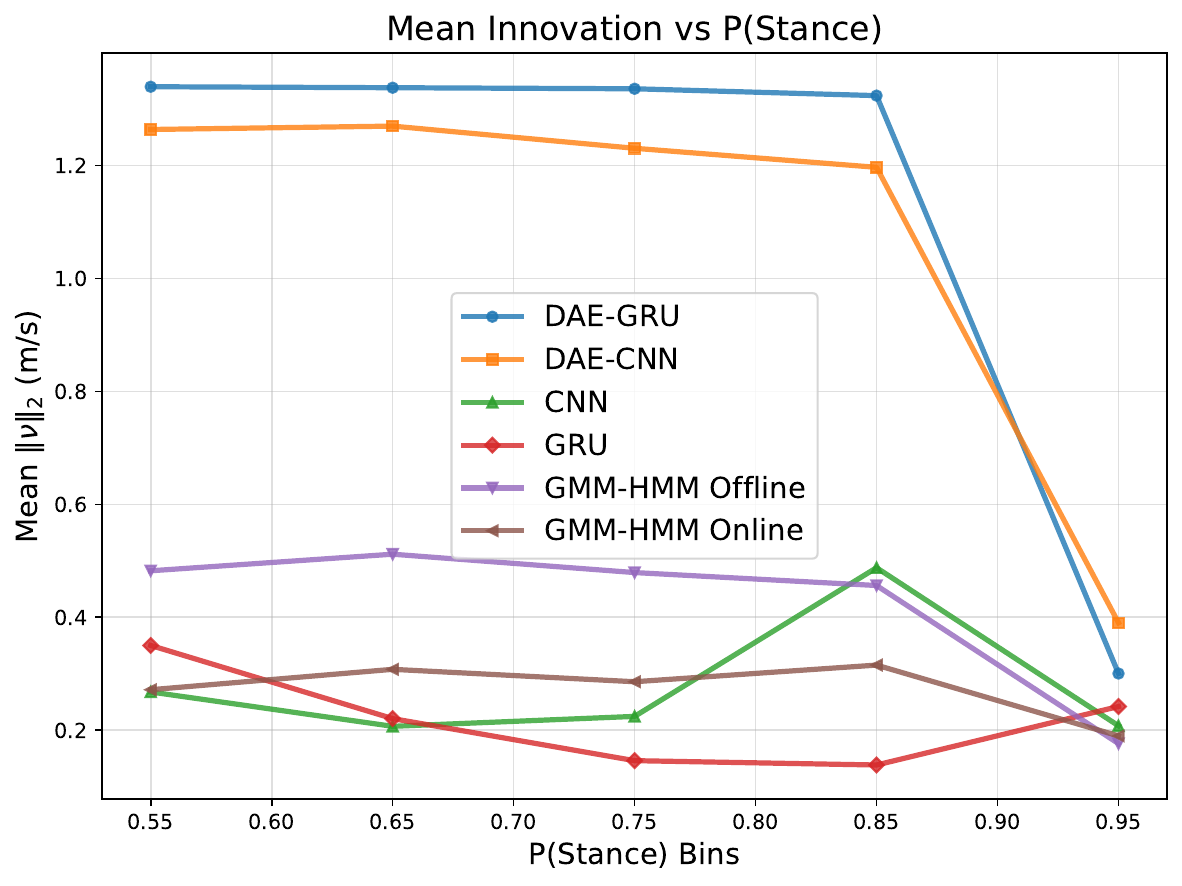}
        \caption{Mean ZUPT innovation norm per stance-score bin on the real-world dataset. For a reliable score, the innovation is high at low scores and decreases towards zero as the score increases.}
        \label{fig:innovation}
    \end{minipage}
\end{figure}

The stance posterior scales the ZUPT measurement covariance (Section \ref{met:contact:hmmgmm}), so we assess whether it behaves as a calibrated probability. Against pseudo-labels derived from the noisy GRF signal, the reliability table (Table \ref{tab:binned_summary}) shows good agreement only in the lowest ($[0.0, 0.1]$) and highest ($[0.9, 1.0]$) bins, and the Expected Calibration Error (ECE) is $0.080$. The output is therefore not calibrated with respect to the GRF-based pseudo-labels.

As discussed in Section \ref{sec:experiments}, however, the GRF signal is an imperfect and potentially unreliable source of supervision. We therefore also evaluate whether the stance score has the expected relationship with the actual ZUPT innovation in the ESEKF: unreliable contacts should produce lower stance scores and larger innovations, whereas reliable contacts should produce higher stance scores and smaller innovations. Figure \ref{fig:innovation} shows that the DAE-GMM pipelines exhibit a clearer monotonic relationship between stance score and ZUPT innovation than the alternative methods. We quantify this relationship with Spearman's rank correlation between the stance score and the innovation magnitude. DAE-GMM (GRU) ($\rho = -0.408$) and DAE-GMM (CNN) ($\rho = -0.380$) show much stronger negative correlations than the closest competitor, HMM-GMM (Offline) ($\rho = -0.029$), so higher stance scores consistently correspond to smaller measurement innovations. Hence, although the GMM output should not be interpreted as a calibrated probability, these results support its use as an ordinal measurement-reliability score for adaptive ZUPT covariance scaling.

\FloatBarrier
\subsection{Effect of Covariance Scaling and Sensitivity Analysis}
\label{appx:covariance_sensitivity}

\begin{figure}[htbp]
    \centering
    \includegraphics[width=0.85\linewidth]{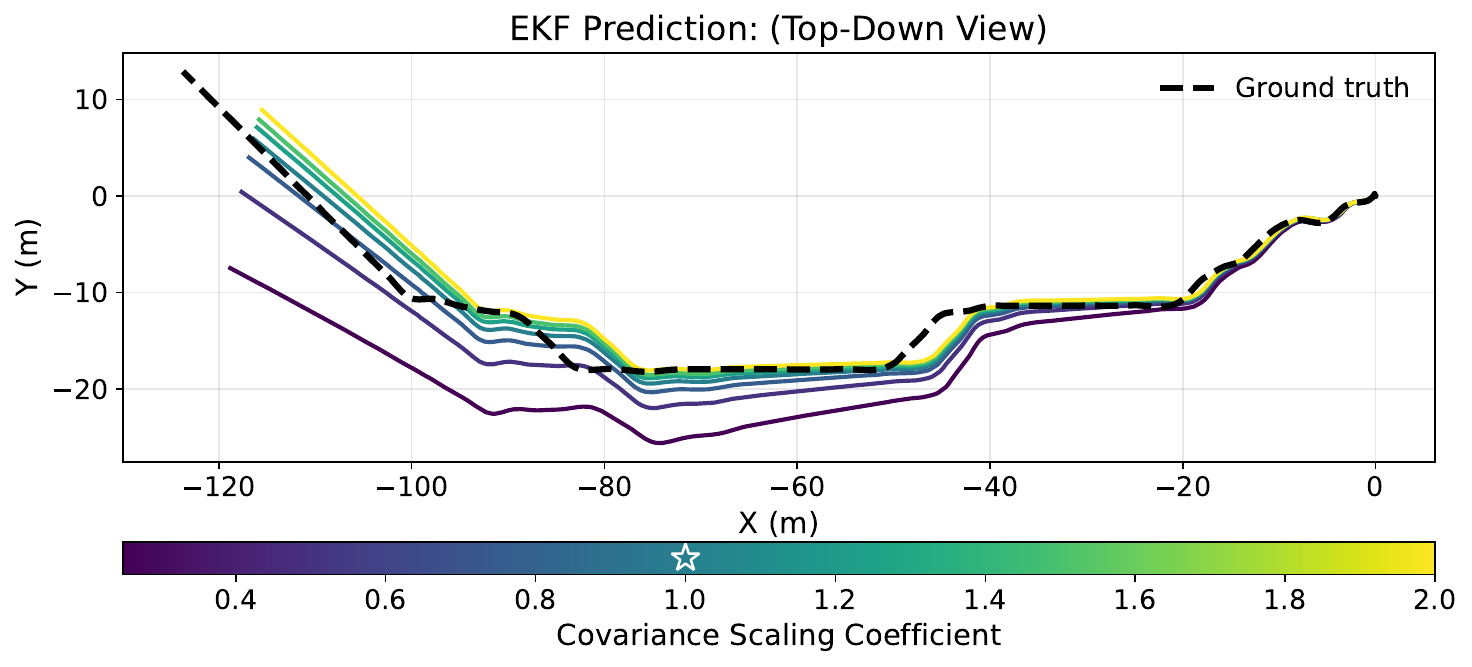}
    \caption{Top-down ESEKF trajectories for different covariance scaling coefficients $\alpha$ (color), compared with the ground truth.}
    \label{fig:covariance}
\end{figure}

The ZUPT measurement covariance is $R_t = \alpha/(p_t+\epsilon)\,I_{3\times3}$ (Section \ref{met:contact:hmmgmm}). The constant $\epsilon=10^{-9}$ is used solely for numerical stability when the reliability score approaches zero and is not a tunable parameter. The overall influence of the adaptive covariance is controlled by the scaling coefficient $\alpha$, which we set to $\alpha=1$ in all experiments to avoid manual tuning. To evaluate the sensitivity to this choice, we run the estimator with $\alpha \in [0.25, 2.0]$ under the same contact states \citep{girgin2026ocelotodometrycontactestimation}, as shown in Figure \ref{fig:covariance}. The results show that $\alpha$ directly affects the yaw drift, which highlights the importance of accurate measurement uncertainty for legged odometry.

\FloatBarrier
\subsection{Effect of the Number of GMM Components}
\label{appx:num_components}

The two-component GMM is not intended as a complete taxonomy of foot-contact conditions, but rather estimates the reliability of the zero-velocity constraint used in the ESEKF. From the perspective of ZUPT, the relevant distinction is whether the foot can be considered stationary, and therefore provides a reliable zero-velocity measurement, or whether the constraint should be weakened or rejected. Slip, partial loading, and transitional contact are physically distinct from swing, but they all reduce the validity of the zero-velocity constraint. In our formulation, such ambiguous conditions are represented by an intermediate stance posterior rather than by explicit additional classes. This is also why the posterior, rather than a hard stance/swing decision, determines the measurement covariance.

\begin{figure}[htbp]
  \centering
  \begin{subfigure}[b]{0.18\textwidth}
        \centering
        \includegraphics[width=\textwidth]{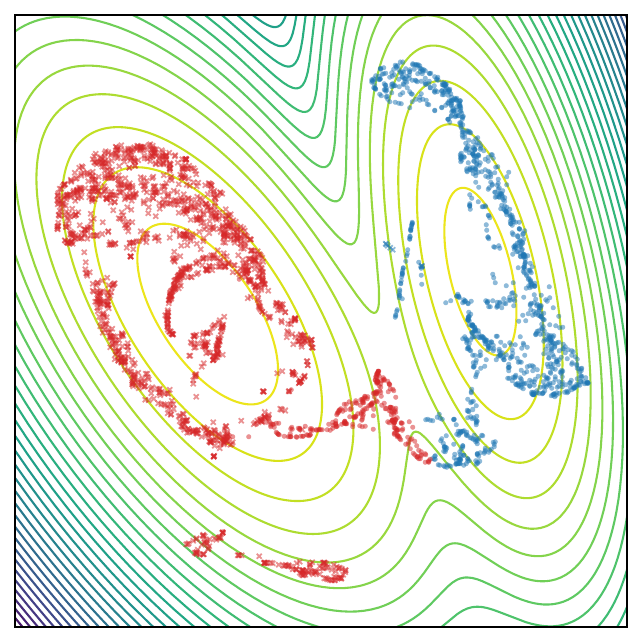}
        \caption{2 components}
    \end{subfigure}
    \hfill
    \begin{subfigure}[b]{0.18\textwidth}
        \centering
        \includegraphics[width=\textwidth]{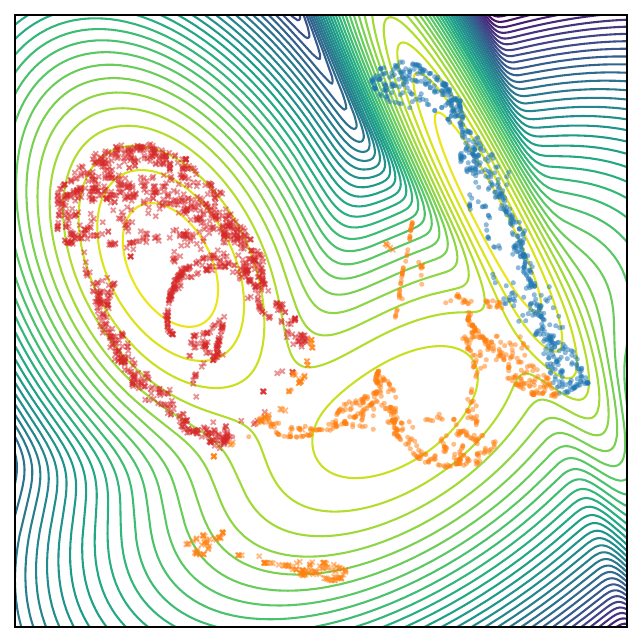}
        \caption{3 components}
    \end{subfigure}
    \hfill
    \begin{subfigure}[b]{0.18\textwidth}
        \centering
        \includegraphics[width=\textwidth]{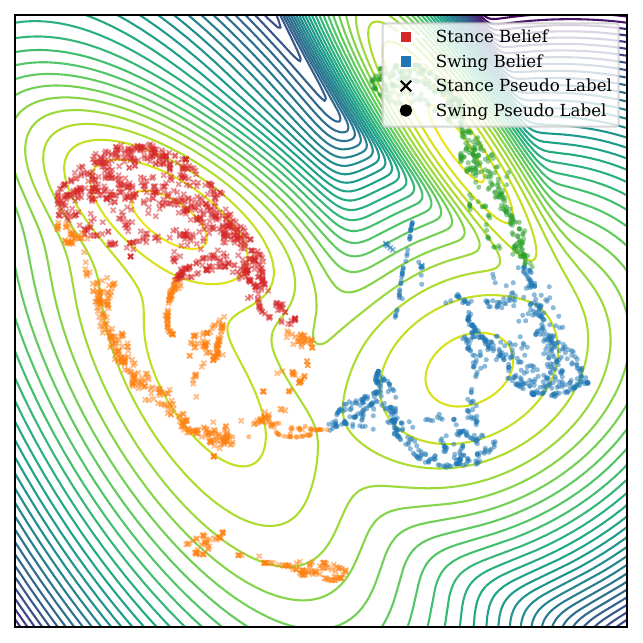}
        \caption{4 components}
    \end{subfigure}
    \hfill
    \begin{subfigure}[b]{0.18\textwidth}
        \centering
        \includegraphics[width=\textwidth]{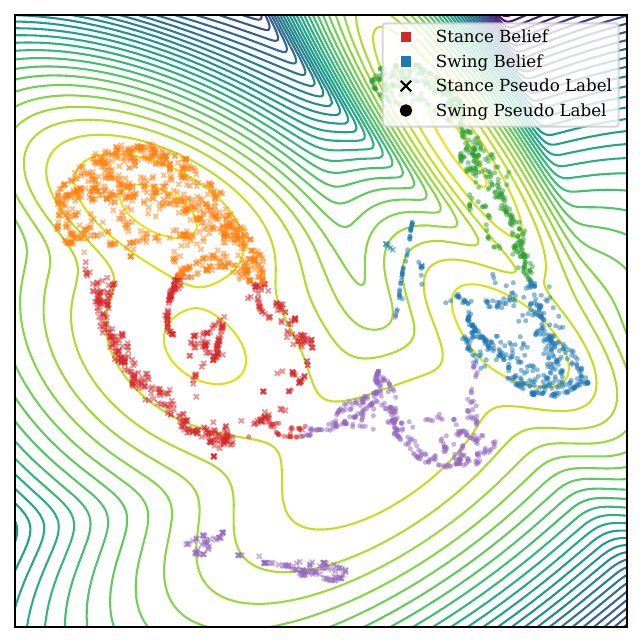}
        \caption{5 components}
    \end{subfigure}
    \hfill
    \begin{subfigure}[b]{0.18\textwidth}
        \centering
        \includegraphics[width=\textwidth]{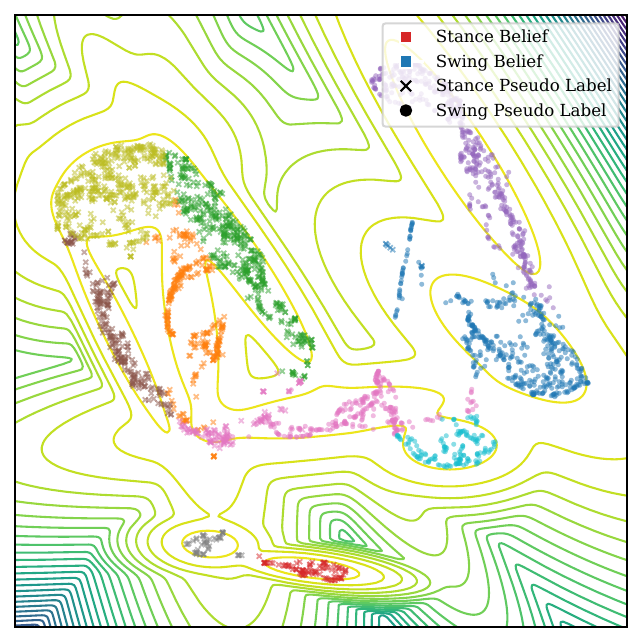}
        \caption{10 components}
    \end{subfigure}
  \caption{UMAP projection of the latent space with GMMs of different numbers of components.}
  \label{fig:n_clusters}
\end{figure}

\begin{figure}[htbp]
    \centering
    \includegraphics[width=0.7\linewidth]{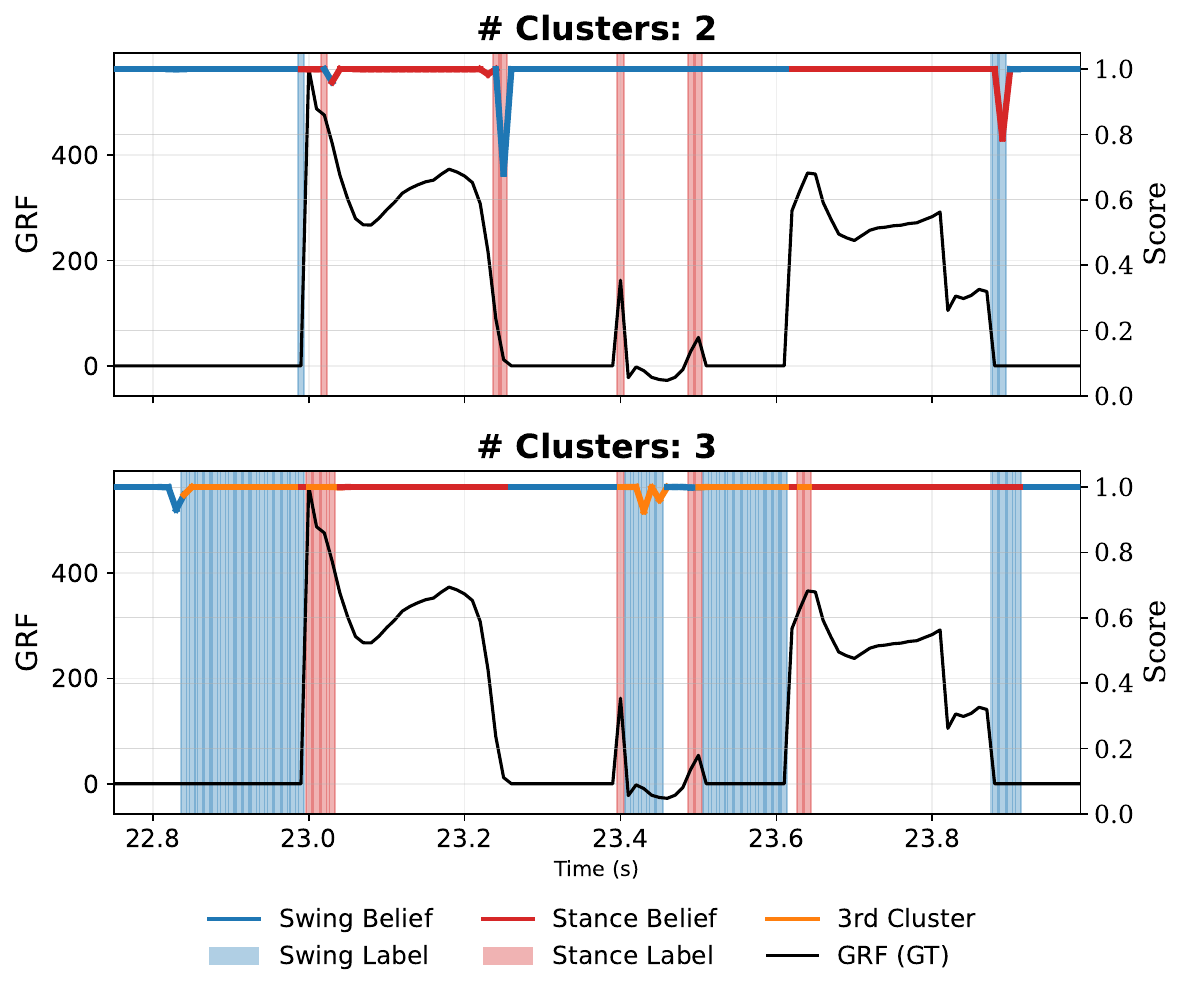}
    \caption{Stance belief over the GRF signal for the 2-component (top) and 3-component (bottom) GMMs. Vertical segments mark disagreements between the belief and the pseudo-label.}
    \label{fig:cluster_vs_grf}
\end{figure}

To investigate whether additional components capture such conditions, we fitted GMMs with 2, 3, 4, 5, and 10 components to the latent space (Figure \ref{fig:n_clusters}). The third component lies in the transition zone between the two dominant modes. Figure \ref{fig:cluster_vs_grf} overlays the resulting beliefs on the GRF signal. Although the third component mostly corresponds to touchdown samples, it does not reveal a consistent pattern such as slip. In this window, a slip occurs between 23.4\,s and 23.5\,s, and the third component does not capture it. Increasing the number of components therefore does not yield a consistent additional contact mode that benefits ZUPT, whereas the two-component model captures the two dominant modes and retains a continuous measure of contact reliability through the stance posterior.

The samples in the overlap region between the two modes (e.g., Figure \ref{fig:window_size_analysis}) are those for which the pseudo-label indicates stance while the posterior indicates swing. As shown in the first row of Figure \ref{fig:cluster_vs_grf}, they mostly correspond to touchdown and liftoff events, as well as slippage.

\FloatBarrier
\subsection{Cluster Assignment under Elevation Changes}
\label{appx:cluster_assignment}

\begin{figure}[htbp]
    \centering
    \begin{minipage}[t]{0.48\linewidth}
        \centering
        \includegraphics[width=\linewidth]{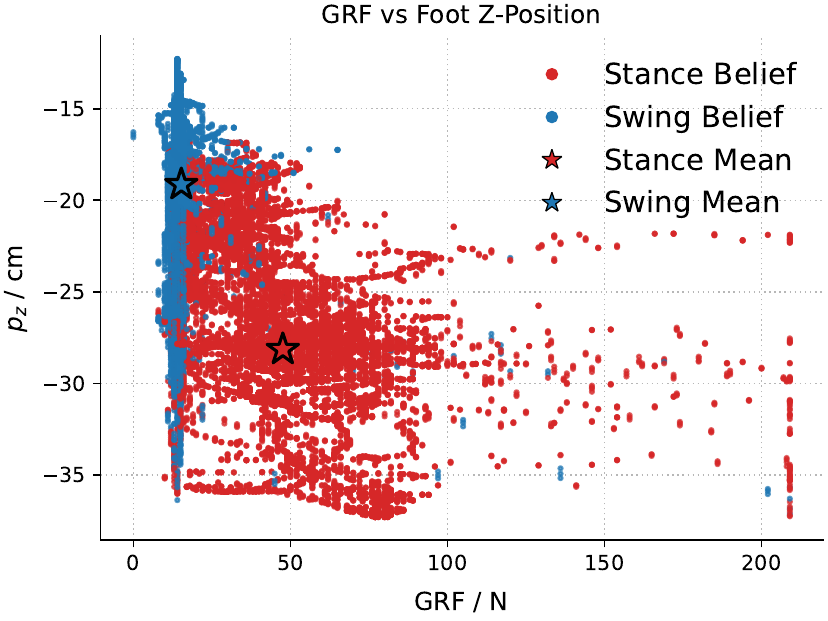}
        \caption{Relative foot z-position vs.\ GRF signal on the staircase sequences, with the stance and swing clusters and their means.}
        \label{fig:foot_position}
    \end{minipage}
    \hfill
    \begin{minipage}[t]{0.48\linewidth}
        \centering
        \includegraphics[width=\linewidth]{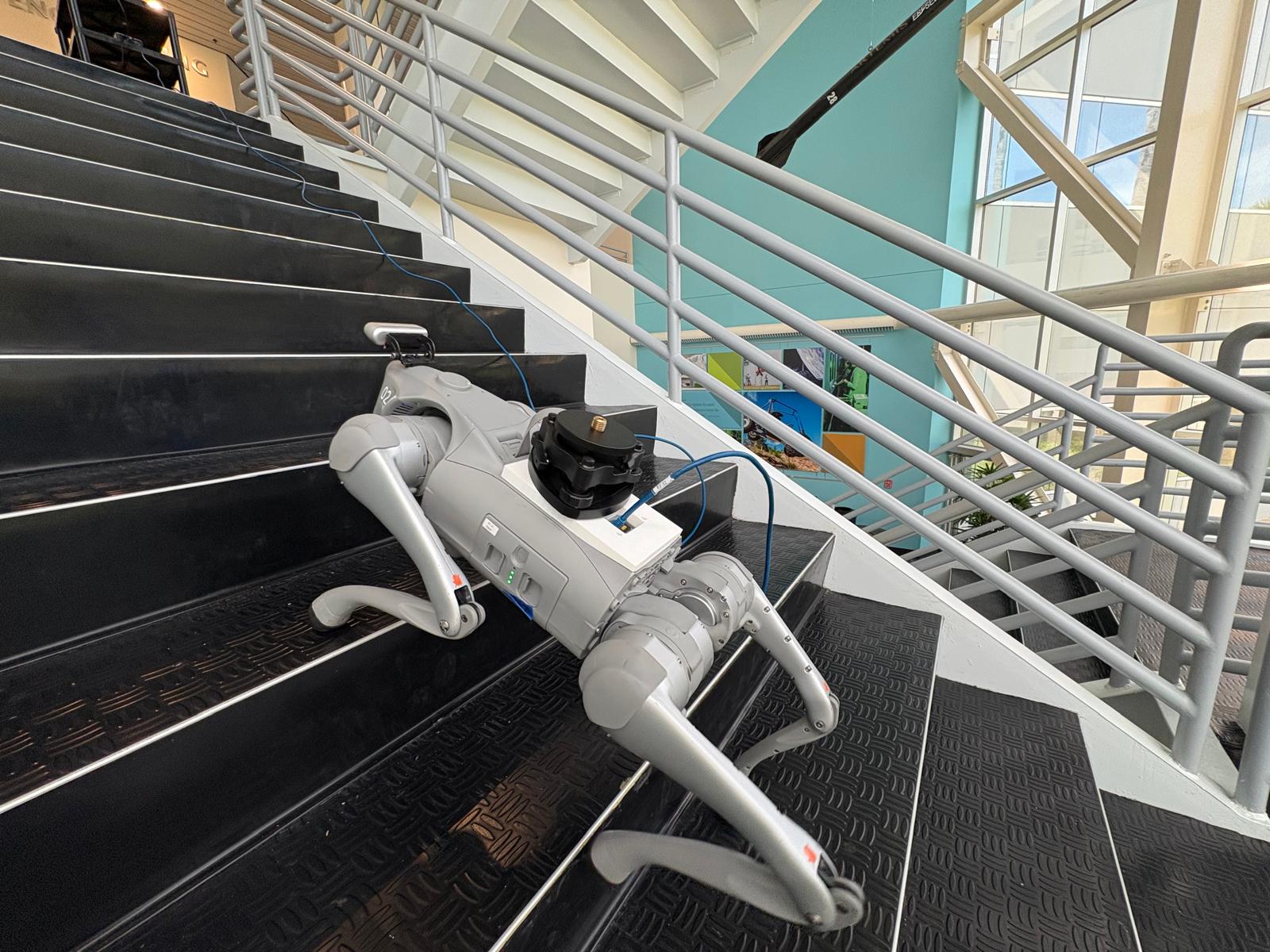}
        \caption{Data collection with the Unitree Go2 on the staircase.}
        \label{fig:unitree_pic}
    \end{minipage}
\end{figure}

The cluster-to-state assignment (Section \ref{met:contact:hmmgmm}) labels the component with the lower mean vertical foot position $p_z^B$ as stance. Because large terrain elevation changes could violate this heuristic, we collected additional real-world sequences during upward and downward staircase traversal with the Unitree Go2 (Figure \ref{fig:unitree_pic}). Figure \ref{fig:foot_position} shows the foot z-position against the GRF signal, together with the color-coded stance and swing clusters.

The staircase introduces substantial variation in foot z-position. Stance samples can occur at relatively high elevations and overlap with swing samples, while swing samples can occur close to the terrain surface. Nevertheless, the two states remain distinguishable, because foot z-position is not used as a threshold on its own but only complements the other features in separating the two GMM clusters. Moreover, the stance cluster still has a lower mean foot z-position than the swing cluster, so the same cluster-to-state assignment remains valid on staircases.

\end{document}